\documentclass{article}

\usepackage{microtype}
\usepackage{graphicx}
\usepackage{subfigure}
\usepackage{booktabs} 
\usepackage{algpseudocode}
\usepackage{multirow}
\usepackage{caption}

\usepackage[hyperfootnotes=false]{hyperref}
\usepackage{xcolor}

\definecolor{tab10_0}{RGB}{31,119,180}
\definecolor{tab10_1}{RGB}{255,127,14}
\definecolor{tab10_2}{RGB}{44,160,44}

\newcommand{\theHalgorithm}{\arabic{algorithm}}

\makeatletter
\providecommand*{\theHALG@line}{\theHalgorithm.\arabic{ALG@line}}
\makeatother

\usepackage[preprint]{icml2026}

\usepackage{amsmath}
\usepackage{amssymb}
\usepackage{mathtools}
\usepackage{amsthm}

\usepackage[capitalize,noabbrev]{cleveref}

\theoremstyle{plain}

\theoremstyle{definition}

\theoremstyle{remark}

\usepackage[textsize=tiny]{todonotes}
\usepackage{booktabs}
\usepackage[normalem]{ulem}
\usepackage{colortbl,array,xcolor}

\usepackage{listings}
\newcommand{\proposed}{\textsc{Mol}LIBRA}
\newcommand{\proposedg}{\textsc{Mol}LIBRA-$\mathcal{G}$}
\newcommand{\proposedl}{\textsc{Mol}LIBRA-$\mathcal{L}$}
\newcommand{\molleo}{\textsc{Mol}LEO}

\newcommand{\labeleddata}{\mathcal{D}_{\mathrm{scored}}}
\newcommand{\unlabeleddata}{\mathcal{D}_{\mathrm{cand}}}
\newcommand{\selecteddata}{\mathcal{D}_{\mathrm{top}}}

\newcommand{\sorteddata}{\mathcal{D}_{\mathrm{sorted}}}
\newcommand{\acqf}{\alpha_{\mathrm{EI}}}
\newcommand{\clamp}{\alpha_{\mathrm{CLAMP}}}
\newcommand{\offspringdata}{\mathcal{D}_{\mathrm{offspring}}}
\newcommand{\elitedata}{\mathcal{D}_{\mathrm{elite}}}
\newcommand{\molenc}{q_\mathrm{mol}}
\newcommand{\moltext}{q_\mathrm{text}}

\newcommand{\argmax}{\mathop{\rm arg~max}\limits}

\newcommand{\funcname}[1]{\textsc{#1}}

\newcommand{\pms}[1]{\ensuremath{\,\text{\scriptsize$\pm$}\,\text{\scriptsize #1}}}
\newcommand{\bestcell}[1]{\cellcolor{gray!30}\underline{\textbf{#1}}}
\newcommand{\secondcell}[1]{\cellcolor{gray!10}\textbf{#1}}

\algnewcommand\algorithmicinput{\textbf{Given:}}
\algnewcommand\Input{\item[\algorithmicinput]}
\algnewcommand\algorithmicoutput{\textbf{Output:}}
\algnewcommand\Output{\item[\algorithmicoutput]}
\begin{document}

\icmltitlerunning{\proposed: Genetic Molecular Optimization with Multi-Fingerprint Surrogates and Text-Molecule Aligned Critic}


\twocolumn[
\icmltitle{\proposed: Genetic Molecular Optimization with \\ Multi-Fingerprint Surrogates and Text-Molecule Aligned Critic}



\icmlsetsymbol{equal}{*}

\begin{icmlauthorlist}
\icmlauthor{Masashi Okada}{pana}
\icmlauthor{Kazuki Sakai}{pana}
\icmlauthor{Hiroaki Yoshida}{pana}
\icmlauthor{Masaki Okoshi}{pana}
\icmlauthor{Tadahiro Taniguchi}{kyoto}
\end{icmlauthorlist}

\icmlaffiliation{pana}{Panasonic Holdings Corp., Japan}
\icmlaffiliation{kyoto}{Kyoto University, Japan}

\icmlcorrespondingauthor{Masashi Okada}{masashi.okada001@jp.panasonic.com}

\icmlkeywords{Machine Learning, ICML}

\vskip 0.3in
]



\printAffiliationsAndNotice{}

\begin{abstract}
We study sample-efficient molecular optimization under a limited budget of oracle evaluations.
We propose \proposed{} (\textbf{M}ultim\textbf{O}da\textbf{L}ity and \textbf{L}anguage \textbf{I}ntegrated \textbf{B}ayesian and evolutiona\textbf{R}y optimiz\textbf{A}tion),
a genetic algorithm based framework that pre-ranks candidate molecules using multiple critics before oracle calls:
(i) an ensemble of Gaussian process (GP) surrogates defined over multiple molecular fingerprints and (ii) a pretrained text-molecule aligned encoder CLAMP.
The GP ensemble enables adaptive selection of task-appropriate fingerprints, %
while CLAMP provides a zero-shot scoring signal from task descriptions by measuring the similarity between molecular and text embeddings. %
On the Practical Molecular Optimization (PMO) benchmark with a budget of 1{,}000 evaluations (PMO-1K), \proposedl{}, our variant with a language-model-based candidate generator, attains the best Top-10 AUC on 14/22 tasks and the highest overall sum of Top-10 AUC across tasks among prior methods.
\end{abstract}

\section{Introduction}
\begin{figure*}
  \centering
  \includegraphics[width=0.85\textwidth]{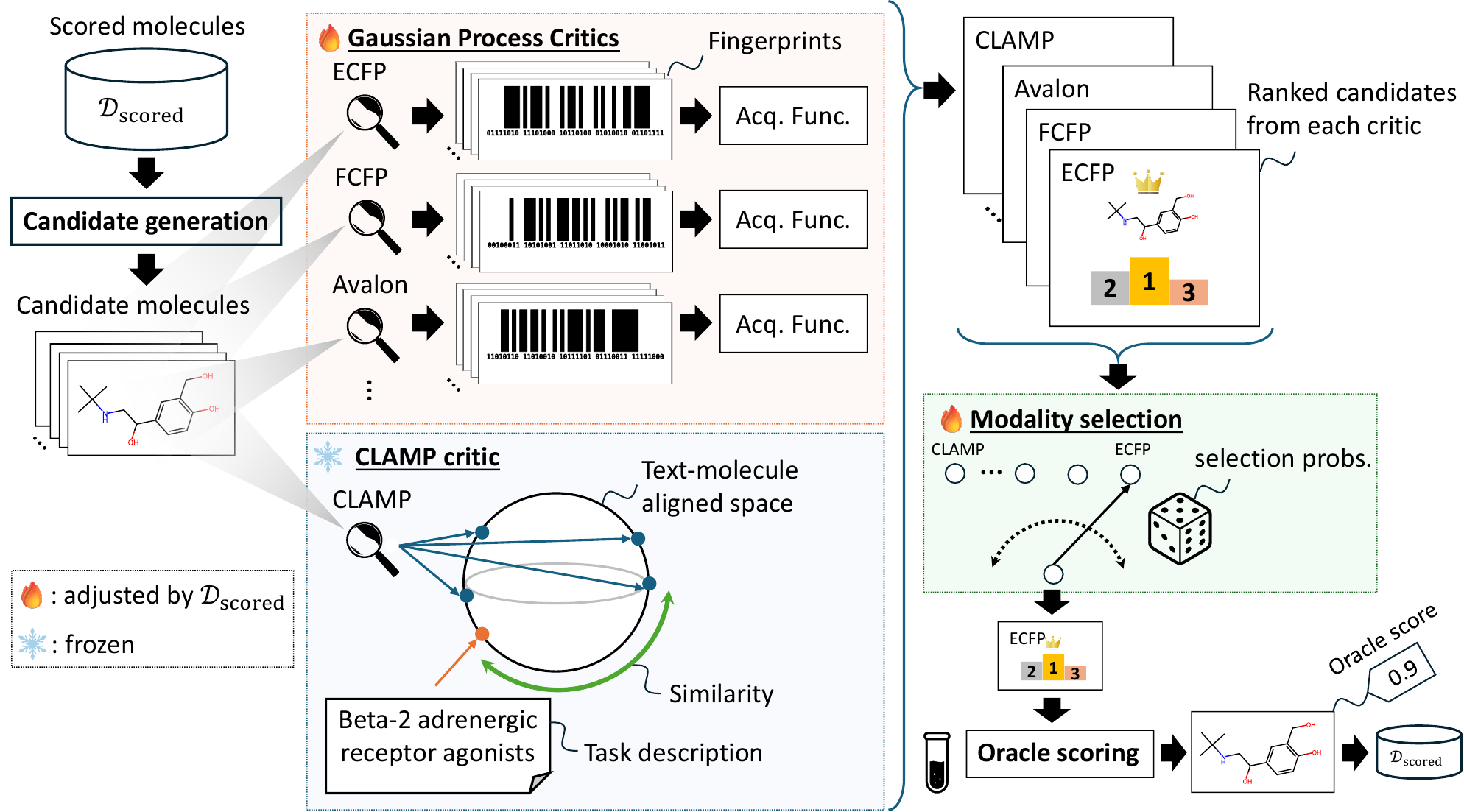}
  \caption{
    A conceptual illustration of \proposed{}, a GA-based molecular optimization framework with multi-fingerprint surrogates and a text--molecule-aligned critic.
    \proposed{} integrates two modalities for pre-evaluation: molecular fingerprints and natural-language task descriptions.
    The critics consist of learnable Gaussian process (GP) models defined over multiple fingerprints and a zero-shot critic based on a pretrained and frozen CLAMP model~\citep{Ramsauer2023CLAMP}.
    A critic is probabilistically selected for candidate ranking, and the selection probabilities are updated using the newly observed oracle scores.
  }
  \label{fig:fig1}
\end{figure*}
Black-box optimization plays an essential role in molecular optimization applications such as drug design, and genetic algorithms (GAs) are a representative approach~\citep{Jensen2019GBGA,moss2020boss,tripp2021fresh,tripp2023genetic,kim2024genetic,Wang2024EfficientEvolutionary,tripp2024diagnosing,yong2025bayesian,lo2025genetic}.
Compared to latent-space optimization methods~\citep{GonzalezDuque2024HDBO,Moss2025COWBOYS,boyar2025conditional} that embed molecules into learned latent spaces using variational autoencoders (VAEs)~\citep{kingma2013auto}, GAs can directly search the space of molecular graphs or strings (e.g., SMILES~\citep{weininger1988smiles}).
In addition, GA operations such as \textit{crossover} and \textit{mutation} can incorporate chemical knowledge through graph-edit rule definitions~\citep{Jensen2019GBGA} or SMILES editing guided by large language models (LLMs)~\citep{Wang2024EfficientEvolutionary}.
In practice, molecular optimization often operates under a limited number of oracle calls, typically on the order of hundreds~\citep{GonzalezDuque2024HDBO,Moss2025COWBOYS,boyar2025conditional}.
This is because evaluating desired properties often requires complex simulations and/or wet-lab experiments, which can be time-consuming and costly~\citep{GarciaOrtegon2022DockString,Stokes2020AntibioticDiscovery}.
However, many molecular optimization methods, including GA-based approaches, assume a large number of oracle calls (e.g., 10,000 calls)~\citep{Jensen2019GBGA,tripp2023genetic,tripp2024diagnosing,xu2024reinvent,Wang2024EfficientEvolutionary,lee2025genmol}, which is impractical in low-budget regimes.

We argue that GA performance in low-budget regimes is limited by the difficulty of constructing reliable surrogate models, due to (i) \textit{fingerprint selection} and (ii) the \textit{cold-start problem}.
In the absence of surrogates to filter candidates~\citep{Jensen2019GBGA,tripp2023genetic,Wang2024EfficientEvolutionary}, GAs may conduct many wasteful evaluations.
In contrast, some GA methods employ Gaussian process (GP) surrogates~\citep{moss2020boss,tripp2023genetic,tripp2024diagnosing,yong2025bayesian}.
To define the GP models, these methods encode molecular structures into vector representations, namely \textit{fingerprints}~\citep{landrum2012fingerprints,adamczyk2024scikit}, to compute similarities between molecules via kernel functions.
Although the optimal fingerprint for accurate prediction depends on the prediction target~\citep{Griffiths2023GAUCHE}, prior GA work typically uses a single fixed fingerprint, which can degrade optimization performance.
Furthermore, GP surrogates require a certain amount of scored data to become reliable, leading to the cold-start problem that wastes precious oracle calls.

To address these challenges, we introduce \proposed{} (\textbf{M}ultim\textbf{O}da\textbf{L}ity and \textbf{L}anguage \textbf{I}ntegrated \textbf{B}ayesian and evolutiona\textbf{R}y optimiz\textbf{A}tion), %
which augments GA search with (i) \emph{multi-fingerprint} GP surrogates and (ii) a \emph{text-molecule aligned critic} based on Contrastive Language-Assay-Molecule Pre-training (CLAMP)~\citep{Ramsauer2023CLAMP} for zero-shot warm-starting (Fig.~\ref{fig:fig1}).
Our main contributions are summarized as follows.
\begin{itemize}
  \item We introduce a \emph{multi-fingerprint} surrogate ensemble: multiple structured-space GP models defined over different fingerprints to reduce sensitivity to fingerprint choice in low budgets.
  \item We integrate a \emph{text-molecule aligned} critic from CLAMP~\citep{Ramsauer2023CLAMP} to provide an early (zero-shot) ranking signal from task descriptions before sufficient scored data are available.
  \item On the Practical Molecular Optimization (PMO) benchmark~\citep{Gao2022SampleEfficiency} under the $N=1{,}000$ budget setting, \proposedl{}, our variant with a language-model-based GA operator, achieves the best Top-10 AUC on 14/22 tasks, and attains the best overall sum of Top-10 AUC across tasks among the compared methods (Table~\ref{tab:top10_auc_full}).
\end{itemize}

The remainder of this paper is organized as follows. 
In Sec.~\ref{sec:preliminaries}, we provide preliminaries on molecular optimization and the building blocks of \proposed{}.
In Sec.~\ref{sec:related_work}, we discuss key differences from related work.
In Sec.~\ref{sec:method}, we present the details of \proposed{}.
In Sec.~\ref{sec:experiments}, we empirically evaluate the performance of \proposed{} on the PMO benchmark.
Finally, Sec.~\ref{sec:discussion} concludes this paper.

\section{Preliminaries} \label{sec:preliminaries}
This section introduces the problem setup and the principal components of \proposed{}.
We first formalize black-box molecular optimization, then review GA-based optimization.
We also summarize GP surrogates, fingerprints, GP ensembles~\citep{lu2023surrogate}, and CLAMP~\citep{Ramsauer2023CLAMP}.

\subsection{Black-box Optimization of Molecular Structures}
The molecular structure optimization problem can be formulated as follows:
\begin{equation}
  x^* = \arg\max_{x \in \mathcal{X}} F(x),
\end{equation}
where $F: \mathcal{X} \to \mathbb{R}$ is an oracle function that evaluates specific properties of molecules (e.g., bioactivity, docking score), $x$ represents a molecular structure (e.g., SMILES, graph), and $\mathcal{X}$ is the set of structures.
The goal is to find an optimal molecule $x^* \in \mathcal{X}$ within a limited oracle call budget $N$.

\subsection{Molecular Optimization via Genetic Algorithm}
The top-level pseudocode of a genetic algorithm (GA) for molecular optimization is presented in Alg.~\ref{alg:molleo}.
This code generalizes existing and proposed GA-based molecular optimization methods by allowing various techniques to be represented through parameter settings and subroutine implementations.

In the main loop of Alg.~\ref{alg:molleo}, the \funcname{GenOffspring} subroutine defined in Alg.~\ref{alg:gen} is called to generate new candidate molecules and add them to the candidate pool $\unlabeleddata$.
Within the \funcname{GenOffspring} subroutine, the \funcname{EditMol} subroutine edits molecular structures in graph or string representations to perform crossover and mutation operations.
GraphGA~\citep{Jensen2019GBGA} implements these editing operations using rule-based methods, whereas \molleo{}~\citep{Wang2024EfficientEvolutionary} requests edits from an LLM.
The pre-evaluation step in Alg.~\ref{alg:molleo} $\ell$9--11 prescreens candidates in $\unlabeleddata$ using a surrogate model $f(x)$ fitted on the scored set $\labeleddata$.
Gaussian process regression, described in the next subsection, is widely used for constructing surrogate models $f(x)$.
In Alg.~\ref{alg:molleo} $\ell$12--16, candidates in $\selecteddata$ are evaluated using the oracle $F$, and the results are added to the observed data $\labeleddata$.
Some of the unevaluated candidates remain in $\unlabeleddata$ and are subject to pre-evaluation again in the next generation ($\ell$17--18).
\algnewcommand{\IIf}[1]{\State\algorithmicif\ #1\ \algorithmicthen}

\begin{algorithm}[t]
\caption{GA-based molecular optimization.}\label{alg:molleo}%
\begin{algorithmic}[1]
\small
\Input{
  Budget $N$, Initial samples $N_{\mathrm{init}}$, Batch size $N_{\mathrm{batch}}$, Oracle function $F$;
  Candidate pool size $N_{\mathrm{cand}}$}
\For {$n \in \{1,\cdots, N_{\mathrm{init}}\}$} \Comment{Initial sampling loop}
  \State $x_{*} \sim p_{\mathrm{init}}(x)$ \Comment{Sample from given molecular dataset}
  \State $\labeleddata \leftarrow \labeleddata \cup (x_{*}, y_{*}=F(x_*))$
\EndFor
\Statex

\State $n \leftarrow N_{\mathrm{init}}$
\State $\unlabeleddata \leftarrow \varnothing$ \Comment{Initialize candidate pool}
\While {True} \Comment{Main loop}
  \State $\unlabeleddata \leftarrow \unlabeleddata \cup \funcname{GenOffspring}(\labeleddata)$
  \State $f \leftarrow \funcname{FitSurrogate}(\labeleddata)$ %
  \State $\sorteddata \leftarrow \funcname{PreEvaluate}(\unlabeleddata, f)$ %
  \Statex \Comment{Sort by pre-evaluated ranking using $f$}
  \State $\selecteddata \leftarrow \sorteddata[1{:}N_{\mathrm{batch}}]$\Comment{Select top-$N_{\mathrm{batch}}$ samples}
  \For {$x_{*} \in \selecteddata$}   \Comment{Oracle call loop} 
    \State $\labeleddata \leftarrow \labeleddata \cup (x_{*}, y_{*}=F(x_{*}))$
    \State $n \leftarrow n + 1$
    \IIf {$n > N$} \textbf{return} $\labeleddata$
  \EndFor
  \State $\sorteddata \leftarrow \sorteddata \setminus \selecteddata$ \Comment{Remove scored samples}
  \State $\unlabeleddata \leftarrow \sorteddata[1{:}N_{\mathrm{cand}}]$  \Comment{Keep candidate pool size}
\EndWhile
\end{algorithmic}
\end{algorithm}

\begin{algorithm}[t]
\caption{
  \funcname{GenOffspring} subroutine. 
}\label{alg:gen}%
\begin{algorithmic}[1]
\small
\Input {Population size $N_\mathrm{elite}$, Number of pairs $N_\mathrm{pairs}$, Number of children (siblings) per pair $N_\mathrm{siblings}$}
\Function{\funcname{GenOffspring}}{$\labeleddata$}
  \State $\offspringdata \leftarrow \varnothing$
  \State $\labeleddata \leftarrow \funcname{SortByScores}(\labeleddata)$
  \Statex \Comment{Sort molecules by their oracle scores in desc.~order}
  \State $\elitedata \leftarrow \labeleddata[1{:}N_\mathrm{elite}]$ 
  \Statex \Comment{Select top-$N_\mathrm{elite}$ molecules}
  \For {$i \in \{1, \cdots, N_\mathrm{pairs}\}$}
    \State $(x, x') \sim \funcname{SampleParents}(\elitedata)$
    \State \{$x_{\mathrm{child}}\} \leftarrow \funcname{EditMol}(x, x', N_\mathrm{siblings})$ 
    \Statex \Comment{Crossover and mutation yielding $N_\mathrm{siblings}$ children}
    \State $\offspringdata \leftarrow \offspringdata \cup \{x_{\mathrm{child}}\}$
  \EndFor
  \State \Return $\offspringdata$ \Comment{Return $N_\mathrm{pairs} \cdot N_\mathrm{siblings}$ molecules}
\EndFunction
\end{algorithmic}
\end{algorithm}

\subsection{Gaussian Process Models}
A Gaussian process (GP) models the posterior distribution of a function $f(x)$ (a surrogate for the oracle $F$) conditioned on observed data $\mathcal{D} = \{(x, y=F(x))\}$;%
\begin{align}
  p(f(x)| \mathcal{D}) = \mathcal{N}(f(x); \mu(x|\mathcal{D}), \sigma^2(x|\mathcal{D})),
\end{align}
where $\mu(\cdot)$ and $\sigma^2(\cdot)$ are the posterior mean and variance defined with a kernel function $k(x,x')$ that measures similarity between $x$ and $x'$.
Commonly used kernels include the Radial Basis Function (RBF) kernel and Mat\'{e}rn kernel defined in continuous spaces; however, applying these kernel functions to molecular structures is infeasible.
Therefore, vector representations of molecules, such as fingerprints, described in the next subsection, are employed.
Hereafter, we refer to GPs based on fingerprints as \textit{structured-space GPs}, following~\citep{Moss2025COWBOYS}.
In prior molecular optimization methods, acquisition functions such as Expected Improvement (EI) and Upper Confidence Bound (UCB) are computed for each candidate molecule using surrogate models to select candidates.
In this study, we employ EI, defined as follows:
\begin{align}
  \acqf(x|\mu,\sigma^{2}) &\triangleq \mathbb{E}_{y \sim \mathcal{N}(y;\mu(x|\cdot), \sigma^{2}(x|\cdot))}\left[\max\left(0, y - y^*\right)\right],
\end{align}
where $y^* = \max_{y \in \mathcal{D}} y$.
\subsection{Fingerprints and Their Similarity Measure}
Molecular fingerprints are fixed-length vectors $x_f \in \mathbb{Z}^{d}$ derived from molecular structures that are directly input to machine learning models, including GPs.
Each dimension of $x_{f}$ represents the presence of specific substructures or chemical groups in the structure $x$ as binary or count values.
Different types of fingerprints can be obtained for the same molecular structure $x$ depending on how substructures are defined~\citep{landrum2012fingerprints,adamczyk2024scikit}, which represent the same molecule through different \textit{lenses}.
Prior molecular optimization studies~\citep{tripp2023genetic,tripp2024diagnosing,Moss2025COWBOYS,yong2025bayesian} conventionally use Extended-Connectivity Fingerprints (ECFP)~\citep{rogers2010extended}, while evaluations with other fingerprints or combinations of multiple fingerprints remain underexplored in the molecular optimization literature.

As a kernel function measuring similarity between molecular fingerprints,
Tanimoto similarity is widely used~\citep{Griffiths2023GAUCHE,tripp2024diagnosing,Moss2025COWBOYS,yong2025bayesian,nguyen2024lico}:
%
\begin{align}
  k(x_{f}, x_{f}') = \frac{x_{f}\cdot x_{f}'}{\|x_{f}\|^2 + \|x_{f}'\|^2 - x_{f}\cdot x_{f}'}, \label{eqn:tanimoto_kernel}
\end{align}
which is equivalent to the well-known Jaccard index, representing the ratio of the intersection to the union.

\subsection{Ensemble of GP Models}
\citep{lu2023surrogate} proposed an ensemble of GP models to improve robustness to kernel selection in Bayesian optimization.
In this method, $M$ models with different kernels $\{p(f_{m}|\cdot) = \mathcal{N}(f_{m};\mu_{m}, \sigma_{m}^2)\}_{m=1}^{M}$ are constructed, each assigned a selection probability $w_{m}$.
This selection probability is updated based on the likelihood of each model whenever a new observation $(x_{*}, y_{*}) \notin \mathcal{D}$ is obtained:
\begin{align}
  w_{m} \leftarrow w_{m} \cdot \mathcal{N}(y_{*}; \mu_{m}(x_{*}|\mathcal{D}), \sigma^2_{m}(x_{*}|\mathcal{D})) \cdot Z^{-1}, \label{eqn:kernel_weight_update}
\end{align}
where $Z$ is a normalization constant ensuring $\sum w_{m} = 1$.
During the computation of the acquisition function, a single model is selected based on these selection probabilities. 
This approach enables online selection of kernels suitable for a current task. 

\subsection{CLAMP}
CLAMP (Contrastive Language-Assay-Molecule Pre-training)~\citep{Ramsauer2023CLAMP} applies the concept of CLIP~\citep{radford2021learning}, a well-known image-text aligned embedding model, to cheminformatics through contrastive pre-training of molecule $x \in \mathcal{X}$ and assay text $a \in \mathcal{S}$ pairs, where $\mathcal{S}$ is the set of natural language strings.
CLAMP employs a molecule encoder $\molenc$: $\mathcal{X} \rightarrow \mathbb{R}^{768}$ and a text encoder $\moltext$: $\mathcal{S} \rightarrow \mathbb{R}^{768}$, trained so that active molecule-text pairs yield high similarity scores $\molenc(x) \cdot \moltext(a) \in \mathbb{R}$.
Pretrained on massive molecular and bioassay text data from PubChem~\citep{Kim2023pubchem}, CLAMP can score molecules for unseen assays described in natural language, enhancing zero-shot activity prediction performance.

\section{Related Work} \label{sec:related_work}
The most relevant works are Tripp's GP Bayesian Optimization (BO)~\citep{tripp2021fresh} and its improved version~\citep{tripp2024diagnosing}.
\citep{tripp2021fresh} has proposed a combination of GraphGA~\citep{Jensen2019GBGA} and a structured-space GP surrogate, and~\citep{tripp2024diagnosing} has reported performance improvements through the appropriate selection of fingerprints and scaling of kernel functions.
\proposed{} builds on this line of work by using diverse fingerprints and incorporating a zero-shot text-molecule aligned critic.

COWBOYS~\citep{Moss2025COWBOYS} is a different type of molecular optimization method that uses structured-space GPs as surrogates, while candidate generation is conducted in latent spaces obtained via VAEs~\citep{kingma2013auto}.
BOSS~\citep{moss2020boss} also combines GA and GP, but the GP is based on SMILES strings, and candidates are generated through string-level GA operations.
GNN-SS and Gradient GA~\citep{wang2023graph,mukherjee2025gradient} construct surrogate models by using graph neural networks (GNNs).
Latent-space Bayesian optimization methods~\citep{GonzalezDuque2024HDBO,Moss2025COWBOYS,boyar2025conditional}, which construct GPs and generate candidates in latent spaces, have been widely studied for molecular optimization.

Recently, methods that use \textit{generative models}, in particular large language models (LLMs), for molecular optimization have gained attention.
\molleo~\citep{Wang2024EfficientEvolutionary}, f-RAG~\citep{lee2024molecule}, and GP-MolFormer-SIM~\citep{navratil2025gp} use LLMs to generate molecular candidates.
\molleo~\citep{Wang2024EfficientEvolutionary} delegates GA's crossover and mutation operations to LLMs.
f-RAG uses retrieval-augmented generation~\citep{lewis2020retrieval}, and GP-MolFormer-SIM~\citep{navratil2025gp} generates molecules through test-time guidance.
Molecular graph generation methods based on diffusion models have also been recently proposed~\citep{liu2025graph,kaech2025refine,lee2025genmol}.
In principle, these generation methods can be combined with the proposed pre-evaluation mechanism; in our experiments, we adopt \molleo{} as a candidate generation engine.
On the other hand, LICO~\citep{nguyen2024lico} uses LLMs as critics, predicting oracle scores through in-context learning~\citep{dong2024survey}.
MT-MOL~\citep{kim2025mtmol} employs LLMs for both generation and pre-evaluation.

Reinforcement learning-based molecular optimization methods~\citep{olivecrona2017molecular,ghugare2023searching,xu2024reinvent,hou2025novo,wang2025leveraging}, represented by REINVENT~\citep{olivecrona2017molecular}, treat molecular generative models as policies and optimize them using oracle scores as rewards.
GFlowNets~\citep{bengio2023gflownet} are methods for sampling molecules from an unnormalized reward-based probability distribution $\propto e^{F(x)}$, and Genetic GFN~\citep{kim2024genetic} is their application to molecular optimization.
MARS~\citep{xie2021mars} is another method for sampling molecules from unnormalized reward-based distributions, employing GNNs and Markov chain Monte Carlo sampling.

\section{Method} \label{sec:method}

\newcommand{\highlightline}[1]{\textcolor{blue}{#1}}

\begin{algorithm}[t]
\caption{Main loop of \proposed{}.} \label{alg:mollibra}%
\begin{algorithmic}[1]
\small
\Input {Budget $N$, Batch size $N_{\mathrm{batch}}$, Candidate pool size $N_{\mathrm{cand}}$, Oracle $F$, \highlightline{CLAMP encoders $\molenc, \moltext$, Task description $a$}}
\State \highlightline{$\{w_m\}_{m=1}^{M} \leftarrow 1/M$} \Comment{Initialize GP weights}
\State \highlightline{$\clamp(x) \triangleq \molenc(x) \cdot \moltext(a)$}
\Statex \Comment{Define zero-shot CLAMP critic}
\While {True} \Comment{Main loop}
  \State $\unlabeleddata \leftarrow \unlabeleddata \cup \funcname{GenOffspring}(\labeleddata)$
  \State \highlightline{$\mathcal{F} \leftarrow \funcname{FitSurrogate}(\labeleddata)$} 
  \Statex \Comment{$\mathcal{F}$: a set of GPs over multiple fingerprints}
  \State \highlightline{$\rho \leftarrow \funcname{Corr}(\{y\}_{y \in \labeleddata} , \{\clamp(x)\}_{x \in \labeleddata})$}   
  \Statex \Comment{Calc.~Spearman corr.~b/w oracle and CLAMP scores}
  \State \highlightline{$\sorteddata \leftarrow \funcname{PreEvaluate}($} %
  \Statex \hspace*{30mm}\highlightline{$\unlabeleddata, \clamp, \rho, \mathcal{F}, \{w_m\})$}
  \State $\selecteddata \leftarrow \sorteddata[1{:}N_{\mathrm{batch}}]$
  \For {$x_{*} \in \selecteddata$}   \Comment{Oracle call loop} 
    \State $\labeleddata \leftarrow \labeleddata \cup (x_{*}, y_{*}=F(x_{*}))$
    \State $n \leftarrow n + 1$
    \IIf {$n > N$} \textbf{return} $\labeleddata$
  \EndFor
  \State \highlightline{$\{w_{m}\} \leftarrow \funcname{UpdateGPWeights}(\labeleddata, \mathcal{F})$} \Comment{Eq.~(\ref{eqn:kernel_weight_update})}
  \State $\sorteddata \leftarrow \sorteddata \setminus \selecteddata$
  \State $\unlabeleddata \leftarrow \sorteddata[1{:}N_{\mathrm{cand}}]$ 
\EndWhile
\end{algorithmic}
\end{algorithm}

\begin{algorithm}[t]
\caption{
  \funcname{FitSurrogate} subroutine in \proposed{}. 
}\label{alg:fit_surrogate}%
\begin{algorithmic}[1]
\small
\Input {Fingerprint types $\{m\}^{M}_{m=1}$}
\Function{\funcname{FitSurrogate}}{$\labeleddata$}
  \State $\mathcal{F} \leftarrow \varnothing$
  \For {Each fingerprint type $m=1$ to $M$}
    \State $\labeleddata^{(m)} \leftarrow \funcname{GetFingerprint}(\labeleddata, m)$
    \State $(\mu_{m}, \sigma^{2}_{m}) \leftarrow \funcname{FitGP}(\labeleddata^{(m)})$
    \State $\mathcal{F} \leftarrow \mathcal{F} \cup \{(\mu_{m}, \sigma^{2}_{m})\}$
  \EndFor
  \State \Return $\mathcal{F}$
\EndFunction
\end{algorithmic}
\end{algorithm}

\begin{algorithm}[t]
\caption{\funcname{PreEvaluate} subroutine in \proposed{}.}\label{alg:pop}%
\begin{algorithmic}[1]
\small
\Function{\funcname{PreEvaluate}}{$\unlabeleddata, \clamp, \rho, \mathcal{F}, \{w_{m}\}$}
  \State $\sorteddata \leftarrow \varnothing$
  \While {$\unlabeleddata \neq \varnothing$} \Comment{While candidates remain}
    \State $u \sim U(0, 1)$  \Comment{Uniform random number}
    \If{$u < \funcname{Clip}(\rho, 0, 1)$}
      \State $x_{*} \leftarrow \argmax_{x \in \unlabeleddata}\ \clamp(x)$    
    \Else
      \State $m \sim \mathrm{Cat}(\{w_{m}\})$ \Comment{Select GP index}
      \State $\unlabeleddata^{(m)} \leftarrow \funcname{GetFingerprint}(\unlabeleddata, m)$
      \State $x_{*} \leftarrow \argmax_{x_{f}^{(m)} \in \unlabeleddata^{(m)}}\ \acqf(x_{f}^{(m)}|\mathcal{F}[m])$
    \EndIf
    \State $\unlabeleddata \leftarrow \unlabeleddata \setminus \{x_{*}\}$ \Comment{Remove selected sample}
    \State $\sorteddata \leftarrow \sorteddata \cup \{x_{*}\}$
  \EndWhile
  \State \Return $\sorteddata$
\EndFunction
\end{algorithmic}
\end{algorithm}

\begin{table*}[t]
\caption{Top-10 AUC across methods on the PMO-1K benchmark (budget $N=1{,}000$). We report mean $\pm$ standard deviation over $n=5$ seeds. For each task, the best and second-best scores are highlighted in \bestcell{bold} and \secondcell{underline}, respectively.}
\label{tab:top10_auc_full}
\resizebox{\textwidth}{!}{%
\begin{tabular}{lcccccccc}
\toprule
& Graph GA & REINVENT & LICO & Genetic GFN & \molleo & Tripp's GP BO & \proposedg & \proposedl \\
& \scriptsize{\citep{Jensen2019GBGA}} & \scriptsize{\citep{olivecrona2017molecular}} & \scriptsize{\citep{nguyen2024lico}} & \scriptsize{\citep{kim2024genetic}} & \scriptsize{\citep{Wang2024EfficientEvolutionary}} & \scriptsize{\citep{tripp2024diagnosing}} & \scriptsize{(ours)} & \scriptsize{(ours)}  \\
\midrule
albuterol\_similarity & 0.583 \pms{0.065} & 0.496 \pms{0.020} & 0.656 \pms{0.125} & 0.664 \pms{0.054} & 0.770 \pms{0.097} & 0.847 \pms{0.031} & \secondcell{0.904 \pms{0.016}} & \bestcell{0.974 \pms{0.007}} \\
amlodipine\_mpo & 0.501 \pms{0.016} & 0.472 \pms{0.008} & 0.541 \pms{0.026} & 0.534 \pms{0.019} & 0.529 \pms{0.023} & 0.555 \pms{0.053} & \secondcell{0.578 \pms{0.044}} & \bestcell{0.619 \pms{0.058}} \\
celecoxib\_rediscovery & 0.424 \pms{0.049} & 0.370 \pms{0.029} & 0.447 \pms{0.073} & 0.447 \pms{0.028} & 0.365 \pms{0.043} & \secondcell{0.471 \pms{0.073}} & 0.418 \pms{0.029} & \bestcell{0.709 \pms{0.068}} \\
deco\_hop & 0.581 \pms{0.006} & 0.572 \pms{0.006} & 0.596 \pms{0.010} & 0.604 \pms{0.017} & 0.603 \pms{0.013} & 0.605 \pms{0.003} & \bestcell{0.635 \pms{0.009}} & \secondcell{0.633 \pms{0.020}} \\
drd2\_docking & 0.833 \pms{0.065} & 0.775 \pms{0.086} & 0.859 \pms{0.066} & 0.809 \pms{0.045} & \secondcell{0.951 \pms{0.010}} & 0.551 \pms{0.477} & 0.867 \pms{0.167} & \bestcell{0.967 \pms{0.010}} \\
fexofenadine\_mpo & 0.666 \pms{0.009} & 0.650 \pms{0.007} & 0.700 \pms{0.023} & 0.682 \pms{0.021} & 0.714 \pms{0.009} & \secondcell{0.755 \pms{0.026}} & \bestcell{0.782 \pms{0.011}} & 0.741 \pms{0.032} \\
gsk3\_beta & 0.523 \pms{0.047} & 0.589 \pms{0.063} & 0.617 \pms{0.063} & \secondcell{0.637 \pms{0.018}} & 0.631 \pms{0.126} & 0.477 \pms{0.092} & 0.520 \pms{0.056} & \bestcell{0.723 \pms{0.067}} \\
isomer\_c7h8n2o2 & 0.735 \pms{0.112} & 0.725 \pms{0.064} & 0.779 \pms{0.099} & 0.738 \pms{0.039} & 0.603 \pms{0.117} & \secondcell{0.890 \pms{0.036}} & \bestcell{0.905 \pms{0.034}} & 0.841 \pms{0.030} \\
isomer\_c9h10n2o2pf2cl & 0.630 \pms{0.086} & 0.630 \pms{0.032} & 0.672 \pms{0.075} & 0.656 \pms{0.075} & 0.601 \pms{0.161} & \secondcell{0.819 \pms{0.059}} & 0.818 \pms{0.053} & \bestcell{0.828 \pms{0.034}} \\
jnk3 & 0.301 \pms{0.071} & 0.315 \pms{0.042} & 0.336 \pms{0.051} & \secondcell{0.409 \pms{0.165}} & 0.359 \pms{0.106} & 0.398 \pms{0.175} & 0.376 \pms{0.156} & \bestcell{0.495 \pms{0.065}} \\
median\_1 & 0.208 \pms{0.015} & 0.205 \pms{0.012} & 0.217 \pms{0.019} & 0.219 \pms{0.008} & 0.157 \pms{0.015} & \secondcell{0.331 \pms{0.009}} & \bestcell{0.335 \pms{0.010}} & 0.263 \pms{0.037} \\
median\_2 & 0.181 \pms{0.009} & 0.188 \pms{0.010} & 0.193 \pms{0.009} & 0.204 \pms{0.011} & 0.207 \pms{0.009} & \secondcell{0.225 \pms{0.022}} & 0.222 \pms{0.022} & \bestcell{0.282 \pms{0.015}} \\
mestranol\_similarity & 0.362 \pms{0.017} & 0.379 \pms{0.026} & 0.423 \pms{0.016} & 0.414 \pms{0.022} & 0.389 \pms{0.034} & 0.556 \pms{0.045} & \bestcell{0.730 \pms{0.125}} & \secondcell{0.630 \pms{0.150}} \\
osimetrinib\_mpo & 0.751 \pms{0.005} & 0.737 \pms{0.007} & 0.759 \pms{0.008} & 0.763 \pms{0.008} & \secondcell{0.795 \pms{0.010}} & 0.748 \pms{0.045} & 0.765 \pms{0.014} & \bestcell{0.828 \pms{0.021}} \\
perindopril\_mpo & 0.435 \pms{0.016} & 0.404 \pms{0.009} & 0.473 \pms{0.030} & 0.462 \pms{0.033} & 0.469 \pms{0.013} & 0.516 \pms{0.037} & \bestcell{0.527 \pms{0.025}} & \secondcell{0.526 \pms{0.019}} \\
rdkit\_qed & 0.914 \pms{0.007} & 0.921 \pms{0.002} & 0.925 \pms{0.005} & 0.928 \pms{0.002} & 0.926 \pms{0.009} & \secondcell{0.929 \pms{0.006}} & 0.928 \pms{0.007} & \bestcell{0.932 \pms{0.011}} \\
ranolazine\_mpo & 0.620 \pms{0.014} & 0.574 \pms{0.044} & 0.687 \pms{0.029} & 0.623 \pms{0.022} & 0.747 \pms{0.012} & 0.752 \pms{0.005} & \secondcell{0.757 \pms{0.015}} & \bestcell{0.777 \pms{0.021}} \\
scaffold\_hop & 0.461 \pms{0.008} & 0.447 \pms{0.010} & 0.480 \pms{0.008} & 0.485 \pms{0.015} & 0.491 \pms{0.055} & 0.495 \pms{0.015} & \bestcell{0.555 \pms{0.014}} & \secondcell{0.536 \pms{0.024}} \\
sitagliptin\_mpo & 0.229 \pms{0.053} & 0.261 \pms{0.026} & 0.315 \pms{0.097} & 0.227 \pms{0.041} & 0.258 \pms{0.087} & 0.353 \pms{0.125} & \secondcell{0.424 \pms{0.110}} & \bestcell{0.434 \pms{0.072}} \\
thiothixene\_rediscovery & 0.322 \pms{0.023} & 0.311 \pms{0.021} & 0.343 \pms{0.035} & 0.377 \pms{0.015} & 0.425 \pms{0.063} & 0.418 \pms{0.068} & \secondcell{0.437 \pms{0.072}} & \bestcell{0.615 \pms{0.017}} \\
troglitazone\_rediscovery & 0.267 \pms{0.015} & 0.246 \pms{0.009} & 0.292 \pms{0.028} & 0.277 \pms{0.015} & 0.271 \pms{0.010} & \bestcell{0.459 \pms{0.048}} & \secondcell{0.451 \pms{0.061}} & 0.412 \pms{0.084} \\
zaleplon\_mpo & 0.374 \pms{0.024} & 0.406 \pms{0.017} & 0.404 \pms{0.022} & 0.400 \pms{0.014} & 0.405 \pms{0.018} & 0.419 \pms{0.024} & \secondcell{0.443 \pms{0.031}} & \bestcell{0.445 \pms{0.028}} \\
\midrule
Sum & 10.901 & 10.673 & 11.714 & 11.559 & 11.665 & 12.569 & \secondcell{13.376} & \bestcell{14.208} \\
\bottomrule
\end{tabular}
}
\end{table*}

\begin{table*}[t]
\centering
\setlength{\tabcolsep}{6pt}
\renewcommand{\arraystretch}{1.1}
\caption{
  Ablation study of \proposedg{} and \proposedl{} under a budget of $N=300$ evaluations on the PMO benchmark.
  We report Top-10 AUC in a similar format to Table~\ref{tab:top10_auc_full}.
  }
\resizebox{0.8\textwidth}{!}{%
\begin{tabular}{l|cccc|cccc}
\toprule
& \multicolumn{4}{c|}{\proposedg} & \multicolumn{4}{c}{\proposedl} \\
\midrule
Multi-fingerprint & \checkmark & \checkmark &  &  & \checkmark & \checkmark &  &  \\
CLAMP critic & \checkmark &  & \checkmark &  & \checkmark &  & \checkmark &  \\
\midrule
albuterol\_similarity & \secondcell{0.748 \pms{0.021}} & \bestcell{0.757 \pms{0.042}} & 0.632 \pms{0.019} & 0.641 \pms{0.024} & \bestcell{0.890 \pms{0.043}} & \secondcell{0.866 \pms{0.076}} & 0.856 \pms{0.085} & 0.735 \pms{0.132} \\
amlodipine\_mpo & \secondcell{0.488 \pms{0.041}} & \bestcell{0.492 \pms{0.046}} & 0.466 \pms{0.028} & 0.472 \pms{0.033} & 0.508 \pms{0.045} & \bestcell{0.517 \pms{0.031}} & 0.497 \pms{0.055} & \secondcell{0.514 \pms{0.035}} \\
celecoxib\_rediscovery & 0.366 \pms{0.026} & \secondcell{0.376 \pms{0.019}} & 0.367 \pms{0.018} & \bestcell{0.381 \pms{0.028}} & \secondcell{0.588 \pms{0.073}} & \bestcell{0.615 \pms{0.079}} & 0.466 \pms{0.082} & 0.499 \pms{0.063} \\
deco\_hop & \secondcell{0.591 \pms{0.010}} & \bestcell{0.600 \pms{0.014}} & 0.588 \pms{0.011} & 0.589 \pms{0.005} & \secondcell{0.597 \pms{0.012}} & \bestcell{0.599 \pms{0.015}} & 0.591 \pms{0.015} & 0.585 \pms{0.005} \\
drd2\_docking & \bestcell{0.669 \pms{0.294}} & 0.485 \pms{0.338} & \secondcell{0.648 \pms{0.298}} & 0.421 \pms{0.367} & \secondcell{0.912 \pms{0.029}} & 0.862 \pms{0.033} & \bestcell{0.914 \pms{0.012}} & 0.724 \pms{0.108} \\
fexofenadine\_mpo & \bestcell{0.689 \pms{0.039}} & \secondcell{0.658 \pms{0.069}} & 0.625 \pms{0.057} & 0.649 \pms{0.061} & \bestcell{0.689 \pms{0.040}} & 0.664 \pms{0.035} & 0.663 \pms{0.059} & \secondcell{0.668 \pms{0.070}} \\
gsk3\_beta & 0.333 \pms{0.054} & \secondcell{0.361 \pms{0.059}} & \bestcell{0.365 \pms{0.052}} & 0.319 \pms{0.077} & \secondcell{0.564 \pms{0.050}} & 0.463 \pms{0.084} & \bestcell{0.626 \pms{0.125}} & 0.505 \pms{0.062} \\
isomer\_c7h8n2o2 & \secondcell{0.751 \pms{0.015}} & 0.745 \pms{0.063} & 0.691 \pms{0.042} & \bestcell{0.764 \pms{0.022}} & \bestcell{0.713 \pms{0.069}} & \secondcell{0.632 \pms{0.099}} & 0.488 \pms{0.359} & 0.534 \pms{0.274} \\
isomer\_c9h10n2o2pf2cl & 0.653 \pms{0.108} & \bestcell{0.696 \pms{0.041}} & 0.625 \pms{0.029} & \secondcell{0.693 \pms{0.079}} & 0.564 \pms{0.270} & \bestcell{0.657 \pms{0.106}} & 0.598 \pms{0.178} & \secondcell{0.646 \pms{0.132}} \\
jnk3 & 0.234 \pms{0.070} & \secondcell{0.244 \pms{0.106}} & 0.211 \pms{0.053} & \bestcell{0.270 \pms{0.130}} & \bestcell{0.297 \pms{0.107}} & 0.263 \pms{0.067} & \secondcell{0.285 \pms{0.052}} & 0.251 \pms{0.097} \\
median\_1 & \secondcell{0.279 \pms{0.011}} & \bestcell{0.287 \pms{0.009}} & 0.270 \pms{0.007} & 0.276 \pms{0.010} & \bestcell{0.225 \pms{0.030}} & \secondcell{0.223 \pms{0.026}} & 0.192 \pms{0.032} & 0.192 \pms{0.026} \\
median\_2 & 0.193 \pms{0.026} & \secondcell{0.194 \pms{0.025}} & 0.191 \pms{0.019} & \bestcell{0.197 \pms{0.021}} & \bestcell{0.230 \pms{0.018}} & 0.222 \pms{0.026} & 0.212 \pms{0.015} & \secondcell{0.230 \pms{0.039}} \\
mestranol\_similarity & \bestcell{0.477 \pms{0.084}} & \secondcell{0.463 \pms{0.045}} & 0.435 \pms{0.041} & 0.432 \pms{0.026} & \bestcell{0.470 \pms{0.020}} & \secondcell{0.456 \pms{0.053}} & 0.430 \pms{0.054} & 0.421 \pms{0.083} \\
osimetrinib\_mpo & \bestcell{0.708 \pms{0.024}} & \secondcell{0.700 \pms{0.027}} & 0.698 \pms{0.040} & 0.700 \pms{0.044} & \bestcell{0.750 \pms{0.039}} & 0.739 \pms{0.050} & 0.736 \pms{0.034} & \secondcell{0.740 \pms{0.020}} \\
perindopril\_mpo & \bestcell{0.455 \pms{0.026}} & \secondcell{0.450 \pms{0.026}} & 0.425 \pms{0.008} & 0.432 \pms{0.029} & \bestcell{0.483 \pms{0.024}} & \secondcell{0.481 \pms{0.020}} & 0.450 \pms{0.045} & 0.455 \pms{0.034} \\
ranolazine\_mpo & \secondcell{0.653 \pms{0.012}} & 0.640 \pms{0.014} & \bestcell{0.669 \pms{0.027}} & 0.651 \pms{0.011} & 0.689 \pms{0.029} & \bestcell{0.709 \pms{0.025}} & 0.695 \pms{0.030} & \secondcell{0.697 \pms{0.029}} \\
rdkit\_qed & 0.903 \pms{0.014} & \bestcell{0.906 \pms{0.013}} & 0.901 \pms{0.009} & \secondcell{0.903 \pms{0.014}} & \bestcell{0.923 \pms{0.006}} & 0.915 \pms{0.012} & 0.917 \pms{0.011} & \secondcell{0.921 \pms{0.005}} \\
scaffold\_hop & \secondcell{0.483 \pms{0.002}} & \bestcell{0.504 \pms{0.020}} & 0.471 \pms{0.011} & 0.473 \pms{0.014} & \bestcell{0.484 \pms{0.013}} & \secondcell{0.482 \pms{0.008}} & 0.458 \pms{0.013} & 0.458 \pms{0.014} \\
sitagliptin\_mpo & \secondcell{0.236 \pms{0.063}} & \bestcell{0.236 \pms{0.067}} & 0.144 \pms{0.090} & 0.174 \pms{0.157} & 0.298 \pms{0.056} & \secondcell{0.300 \pms{0.080}} & 0.287 \pms{0.091} & \bestcell{0.348 \pms{0.053}} \\
thiothixene\_rediscovery & \secondcell{0.326 \pms{0.036}} & \bestcell{0.356 \pms{0.051}} & 0.309 \pms{0.037} & 0.316 \pms{0.044} & \bestcell{0.480 \pms{0.079}} & 0.445 \pms{0.041} & 0.440 \pms{0.044} & \secondcell{0.470 \pms{0.072}} \\
troglitazone\_rediscovery & 0.316 \pms{0.031} & \bestcell{0.335 \pms{0.030}} & 0.319 \pms{0.019} & \secondcell{0.334 \pms{0.018}} & \secondcell{0.310 \pms{0.030}} & \bestcell{0.312 \pms{0.033}} & 0.302 \pms{0.027} & 0.309 \pms{0.036} \\
zaleplon\_mpo & \bestcell{0.386 \pms{0.026}} & \secondcell{0.382 \pms{0.025}} & 0.311 \pms{0.032} & 0.350 \pms{0.040} & 0.394 \pms{0.053} & \secondcell{0.400 \pms{0.037}} & 0.367 \pms{0.055} & \bestcell{0.403 \pms{0.039}} \\
\midrule
Sum & \bestcell{10.934} & \secondcell{10.868} & 10.362 & 10.439 & \bestcell{12.058} & \secondcell{11.823} & 11.469 & 11.306 \\
\bottomrule
\end{tabular}
}
\label{tab:ablation_full}
\end{table*}

This section describes the proposed method, \proposed{}, whose algorithm is presented in Algs.~\ref{alg:mollibra}, \ref{alg:fit_surrogate}, and \ref{alg:pop}.
The top-level description shown in Alg.~\ref{alg:mollibra} is almost the same as Alg.~\ref{alg:molleo}, so we omit the initialization loop and highlight the different lines from Alg.~\ref{alg:molleo}.
A distinctive feature of \proposed{} is the use of multiple surrogates (or critics).
By constructing and ensembling structured-space GPs over multiple fingerprint representations, we enhance robustness against uncertainty in molecular representations.
We adopt an ensemble method from \citep{lu2023surrogate}, which was originally developed for adaptive kernel selection.
Additionally, by incorporating a zero-shot critic using CLAMP, we aim to improve optimization performance in low-budget regimes.

\subsection{Definition of Multiple Critics}
The construction of multiple GP models is performed by the \funcname{FitSurrogate} subroutine shown in Alg.~\ref{alg:fit_surrogate}, where molecules in $\labeleddata$ are represented by $M$ types of fingerprints, and one GP is constructed for each.
The CLAMP critic is defined in Alg.~\ref{alg:mollibra} $\ell$2, where $a$ is a text that describes the desired molecule, such as $a = \textit{``Inhibition of L-type calcium channel''}$.
We use the official CLAMP model and weights%
\footnote{
  \url{https://github.com/ml-jku/clamp/tree/main}.
}.

\subsection{Selection of Multiple Critics' Results}
Multiple critics constructed in the previous section are selected based on probabilities denoted by $\{w_{m}\}_{m=1}^{M}$ and $\rho$.
Here, $\{w_{m}\}_{m=1}^{M}$ are the weights for each GP model, which are initialized uniformly in Alg.~\ref{alg:mollibra} $\ell$1 and then updated based on Eq.~(\ref{eqn:kernel_weight_update}) ($\ell$14) using newly acquired observations.
The value $\rho$ computed in Alg.~\ref{alg:mollibra} $\ell$6 is the Spearman correlation coefficient $\rho \in [-1, 1]$ between the oracle scores and CLAMP critic values in the evaluated data $\labeleddata$.
$\rho$ is used as a gating probability to select between CLAMP and the GP models.
Unlike GPs, CLAMP cannot evaluate the likelihood of new data as in Eq.~(\ref{eqn:kernel_weight_update}); therefore, we use the correlation coefficient instead.

Based on the aforementioned selection probabilities, candidates in $\unlabeleddata$ are ranked and sorted by the \funcname{PreEvaluate} subroutine shown in Alg.~\ref{alg:pop}.
Here, a critic is probabilistically selected for each rank, and the result of that critic is adopted.
We expect that switching critics within the loop will help ensure diversity in the candidate pool $\unlabeleddata$.
The CLAMP critic is selected based on $\rho$ clipped to $[0, 1]$.
If CLAMP is not selected, a GP model is selected based on the probability distribution $\{w_{m}\}_{m=1}^{M}$, and the candidate that maximizes the acquisition function $\acqf(x|\mu_{m}, \sigma^{2}_{m})$ based on that GP is selected.

\subsection{Implementation Details}
\paragraph{Fingerprints, kernels and GP models}
In this paper, we adopt $M=6$ types of fingerprints: ECFP, FCFP (Functional Connectivity Fingerprint)~\citep{rogers2010extended}, Avalon~\citep{gedeck2006qsar}, Pharmacophore~\citep{mcgregor1999pharmacophore}, MAP (MinHashed Atom Pair)~\citep{orsi2024one}, and BoC (Bag-of-Characters)~\citep{Griffiths2023GAUCHE}.
BoC is a vector representation that counts the frequency of occurrence of characters contained in the SMILES string of a molecule; although it is not strictly a fingerprint, we treat it as a type of fingerprint in this paper.
For all these fingerprints, we adopt the Tanimoto kernel defined in Eq.~(\ref{eqn:tanimoto_kernel}).
We perform training and inference of the GP models using BoTorch~\citep{balandat2020botorch}.

\paragraph{Candidate generation}
In this paper, we integrate both GraphGA's graph editing~\citep{Jensen2019GBGA} and \molleo's LLM-based SMILES generation~\citep{Wang2024EfficientEvolutionary} into \proposed{} to realize the \funcname{EditMol} subroutine; these implementations are denoted as \proposedg{} and \proposedl{}, respectively.
The LLM we use is GPT-5-mini~\citep{openai2025gpt5}, and the prompt for molecular editing is shown in Appendix List~\ref{lst:prompt_template}.
Sampling from $p_{\mathrm{init}}(x)$ in the initialization loop is performed by random selection from the ZINC-250K dataset~\citep{gomez2018automatic} as in \molleo{}.
Hyperparameters related to candidate generation are detailed in Appendix Table~\ref{tab:hparams}.

\section{Experiments} \label{sec:experiments}

\subsection{Settings}
\paragraph{Molecular optimization tasks}
We evaluate on 22 selected optimization tasks from the PMO benchmark~\citep{Gao2022SampleEfficiency}%
\footnote{
The PMO benchmark is commonly evaluated on 23 tasks; however, we exclude the \texttt{valsartan\_smarts} task in our experiments.
This task employs a sparse oracle that returns 0 unless a specific substructure \texttt{CN(C=O)Cc1ccc(c2ccccc2)cc1} is present, and many methods have failed to optimize it.
\citep{lo2025genetic} points out that this task is ill-suited for benchmarking, and we follow their recommendation.
}, a standard benchmark suite for molecular optimization.
To specifically evaluate performance under a low oracle budget, the oracle budget for each task is limited to $N=1{,}000$ oracle calls; this setting is referred to as PMO-1K~\citep{nguyen2024lico,kim2025mtmol}.

\paragraph{Task descriptions for LLM and CLAMP}
We classify the tasks into three categories and provide task descriptions to the LLM and CLAMP according to each category.
The categories and examples of task descriptions are shown in Table~\ref{tab:task-objectives-short}.

\begin{table}[t]
  \centering
  \setlength{\tabcolsep}{2pt}
  \caption{
    Task description examples provided to CLAMP.
    One example is given for each of the three task categories.
    The definition rules for task descriptions in this paper and full version of this table are provided in Appendix~\ref{sec:appendix_prompt} and Table~\ref{tab:task-objectives}.
  }
  \label{tab:task-objectives-short}
  \resizebox{\columnwidth}{!}{%
  \begin{tabular}{ll}
    \toprule
    Category & Example of Task description $a$\\
    \midrule
    (i) Drug-Effect Opt. & ``\textit{Beta-2 adrenergic receptor agonists}'' \\
    & (from \texttt{albuterol\_similarity}) \\
    (ii) Property Opt. & ``\textit{Quantitative Estimate of Drug-likeness}'' \\
    & (from \texttt{rdkit\_qed}) \\
    (iii) Structure-Constrained Opt. & `` '' (empty string) (from \texttt{deco\_hop}) \\
    \bottomrule
  \end{tabular}
  }
\end{table}


\paragraph{Baselines}
We compare \proposed-$\{\mathcal{G},\mathcal{L}\}$ against six representative methods in PMO, namely Tripp's GP BO~\citep{tripp2024diagnosing}, \molleo~\citep{Wang2024EfficientEvolutionary}, Genetic GFN~\citep{kim2024genetic}, LICO~\citep{nguyen2024lico}, REINVENT~\citep{olivecrona2017molecular}, and Graph GA~\citep{Jensen2019GBGA}.
We implement Tripp's GP BO and \molleo{} as special cases of \proposed-$\{\mathcal{G},\mathcal{L}\}$.
Tripp's GP BO uses a single fingerprint (ECFP) and disables CLAMP in \proposedg{}, which matches the improved version described in~\citep{tripp2024diagnosing}.
\molleo{} disables all critics in \proposedl{}, using the same prompt template as \proposedl{}.
For the other baselines (i.e., Genetic GFN, LICO, REINVENT, and Graph GA), we cite results from a previous study~\citep{nguyen2024lico}.

\paragraph{Metrics}
We use the Top-10 AUC metric to evaluate optimization performance,
which measures the area under the curve of the average of the top-10 highest oracle scores obtained up to each evaluation step.
This metric reflects both the quality of the best-found molecules and the speed of optimization.

\subsection{Results}
\paragraph{Comparison with the baselines}
The purpose of this experiment is to compare our two variants, \proposedg{} and \proposedl{}, with leading baselines on PMO-1K.
Table~\ref{tab:top10_auc_full} presents the Top-10 AUC scores for each method across tasks.
\proposedg{} and \proposedl{} achieve the best scores in 7 and 14 out of 22 tasks, respectively, showing consistent gains over all baselines.
Tripp's GP BO ranks third overall, outperforming LLM-based methods such as \molleo{} and LICO, which differs from the results reported in~\citep{nguyen2024lico}.
This difference is likely due to~\citep{nguyen2024lico} using an earlier GP BO implementation from~\citep{tripp2021fresh} instead of the improved version from~\citep{tripp2024diagnosing}.

\paragraph{Ablation study}
The purpose of this ablation study is to evaluate the effectiveness and contributions of each component of \proposed{}.
In this experiment, we evaluated three variants that remove either or both of the main components of the proposed method: multi-fingerprint and CLAMP critic.
The variant without multi-fingerprint uses only a single fingerprint (ECFP).
The variant that removes both from \proposedg{} corresponds to Tripp's GP BO.
The trial budget for this evaluation was set to $N=300$ (PMO-300).
Table~\ref{tab:ablation_full} presents the results of the ablation study.
Overall, removing either component can degrade performance on multiple tasks, indicating that both multi-fingerprint and CLAMP can be beneficial under low budgets.

\begin{figure}
  \centering
  \includegraphics[width=0.45\textwidth]{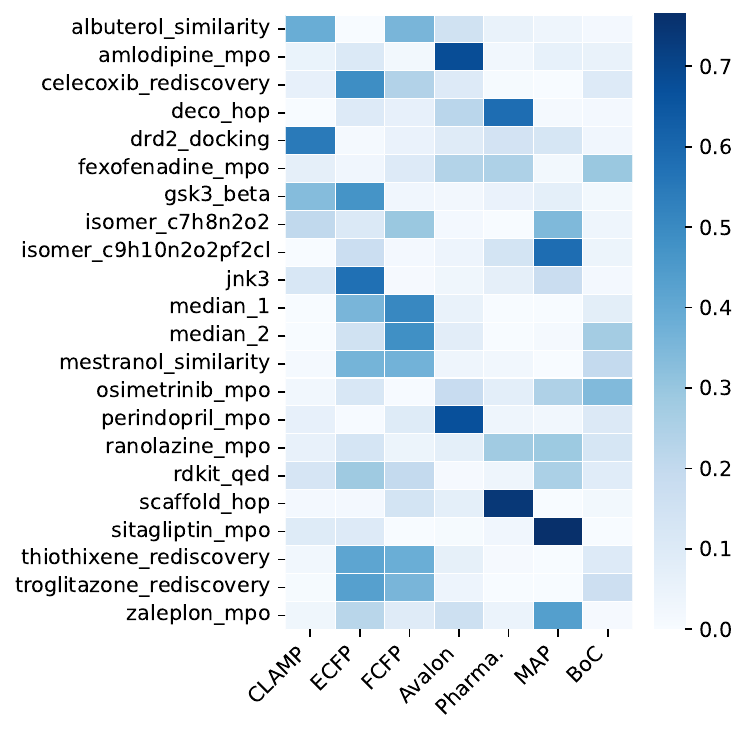}
  \caption{Heatmap visualizing the contribution of critics (structured-space GPs and the CLAMP critic) in \proposedl{}'s optimization process. 
  The color intensity indicates the accumulation of step-wise improvement in oracle scores realized by each critic. 
  In the figure, the contributions are normalized so that the total contribution of all critics sums to 100\%.
  Similar results for \proposedg{} are provided in Appendix Figure~\ref{fig:modal_contrib_mollibrag}.
  }
  \label{fig:modal_contrib_mollibral}
\end{figure}

\paragraph{Contribution analysis of critics}
The purpose of this analysis is to understand the behavior of each critic (structured-space GPs and the CLAMP critic) during optimization and to support the necessity of the proposed adaptive critic selection.
The analysis here focuses on \proposedl{}.
Figure~\ref{fig:modal_contrib_mollibral} visualizes the \textit{contributions} of each critic in the optimization process.
Here, the contribution is measured as the accumulation of step-wise score improvement realized by each critic over the entire process and all seeds.
This result indicates that the contributing fingerprints vary across tasks, and the CLAMP critic also contributes non-trivially in multiple tasks; e.g., \texttt{albuterol\_similarity}, \texttt{drd2\_docking} and \texttt{gsk3\_beta}.

\begin{figure}
  \centering
  \includegraphics[width=0.45\textwidth]{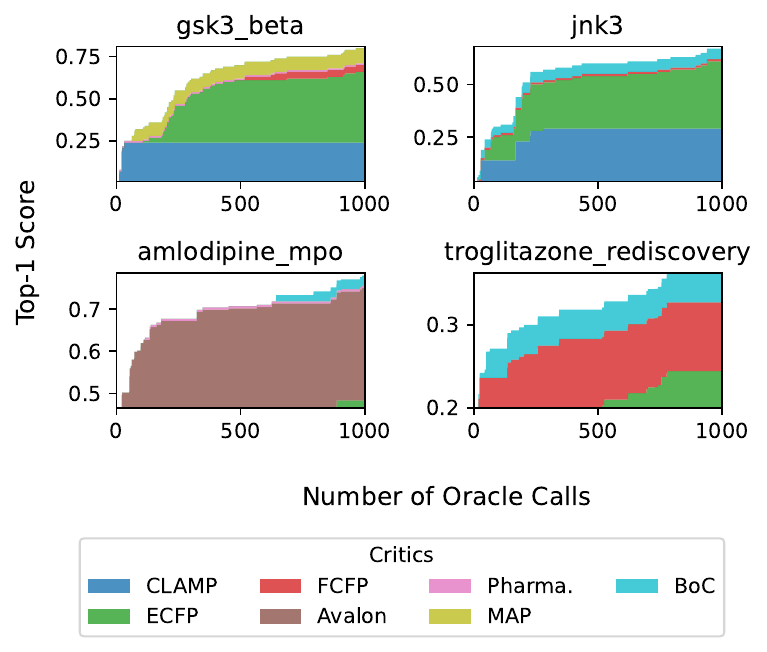}
  \caption{ 
  Temporal evolution of contributions in four tasks (results from a single seed run).
  The cumulative score improvement realized by each critic is shown as an area chart.
  }
  \label{fig:modal_contrib_mollibral_steps}
\end{figure}

Figure~\ref{fig:modal_contrib_mollibral_steps} shows the temporal evolution of contributions in four representative tasks.
In the upper row (\texttt{gsk3\_beta} and \texttt{jnk3}), the CLAMP critic contributes most in the early steps.
In contrast, there are tasks where CLAMP contributes negligibly in our runs (lower row: \texttt{amlodipine\_mpo} and \texttt{troglitazone\_rediscovery}), which we discuss in the next section.
In \texttt{troglitazone\_rediscovery}, FCFP takes over from ECFP as the main contributor around step 500, indicating that the contributions of fingerprints can change during the optimization process.
Moreover, as shown in Appendix Figure~\ref{fig:modal_contrib_mollibral_seeds}, the contributions of critics can vary significantly across different seeds. This suggests that the effective fingerprints depend on the initial distribution of $\labeleddata$.
The above results indicate that it is challenging to select an effective single fingerprint in advance for specific tasks. 
We can therefore conclude that multi-fingerprint ensembling is essential.

\paragraph{Additional experiments}
Appendix~\ref{sec:appendix_additional_results} presents the results of various additional experiments, including extra contribution analyses and comparisons with latent-space optimization methods on PMO-300, which further support the effectiveness of \proposed{}.
We also provide further ablation studies on \proposed{}'s components (i.e., GP ensemble and CLAMP gating method), where alternative components are evaluated to validate the design choices of the proposed method.


\section{Discussion} \label{sec:discussion}
We present \proposed{}, a GA-based molecular optimization method that combines an ensemble of structured-space GPs over multiple molecular fingerprints with a zero-shot CLAMP critic.
Using multiple fingerprints improves robustness to representation choice, and incorporating CLAMP can strengthen performance in low-budget regimes.
To our knowledge, this is the first work that incorporates both
multi-fingerprint structured-space GPs and a CLAMP critic for molecular optimization, and it achieves the best overall Top-10 AUC among the compared methods on PMO-1K (Table~\ref{tab:top10_auc_full}).

A limitation of \proposed{} is that the CLAMP critic is not consistently helpful across tasks.
One possible explanation is that CLAMP's molecular encoder is based on ECFP-derived features~\citep{Ramsauer2023CLAMP}, which can overlap with information captured by ECFP-based GP critics.
Since the most effective fingerprint differs across tasks, an ECFP-based CLAMP critic can be misaligned with the task-specific representation that best supports optimization.

Future work includes extending CLAMP to incorporate multiple fingerprint representations, which may improve its effectiveness as a zero-shot critic.
Another direction is to expand the set of critics, for example by adding fingerprints not used in this study, using GP critics with the DICE kernel~\citep{boyar2025conditional} as an alternative to the Tanimoto kernel, or combining with LLM-based critics such as LICO~\citep{nguyen2024lico}.

\bibliographystyle{icml2026}
\bibliography{references}

\clearpage
\appendix
\section{Experimental Details} \label{sec:appendix_experimental_details}
\subsection{Hyperparameters}
Table~\ref{tab:hparams} summarizes the hyperparameters used in our experiments.
The hyperparameters for \molleo{} and Tripp's GP BO, which are based on the implementation of \proposed{}, are also summarized.
Since \molleo{} does not have a pre-evaluation mechanism, it evaluates all generated molecules with the oracle ($N_\mathrm{batch} = N_\mathrm{pairs} \cdot N_\mathrm{siblings}$) and does not use a candidate pool ($N_\mathrm{cand} = 0$).
Also, considering the limited budget of $N=1{,}000$ evaluations, we set $N_\mathrm{batch} = 10$ in \molleo{} to reduce the number of oracle calls per generation (originally $N_\mathrm{batch} = 70$ in the official code%
\footnote{\url{https://github.com/zoom-wang112358/MOLLEO/tree/main}}).
On the other hand, \proposed{} and Tripp's GP BO utilize the pre-evaluation mechanism: we set $N_\mathrm{pairs}\cdot N_\mathrm{siblings} = 50$ and $N_\mathrm{cand} = 300$ to ensure candidate diversity, and we use $N_\mathrm{batch} = 1$ to increase the number of feedback iterations.

\subsection{Prompt Template and Task Descriptions} \label{sec:appendix_prompt}
List.~\ref{lst:prompt_template} shows the prompt template used for LLM-based molecular editing in \proposedl{}.
We adopt a structured prompt~\citep{berryman2024prompt} based on the original template in the official code of \molleo{}.
We use this prompt template for both \proposedl{} and \molleo{}.

Table~\ref{tab:task-objectives} lists the task classification and task descriptions used in the prompt templates and CLAMP text inputs.
The categories are
(i) drug-effect optimization,
(ii) property optimization, and
(iii) structure-based optimization.
For (i) drug-effect tasks, which are originally synthetic tasks that maximize similarity to known drug molecules, we reinterpret them as searching for molecules with similar effects or activities to known drugs.
In the task descriptions, we only include descriptions of the desired effects or activities%
\footnote{
\molleo{}~\citep{Wang2024EfficientEvolutionary} and MT-MOL~\citep{kim2025mtmol} included specific drug names in their LLM prompts.
}.
For (ii) property optimization tasks, the names of the subject properties are included in the task descriptions.
For (iii), structure-based tasks aim to search for molecules that satisfy structural constraints.
For the structure-based tasks, defining a meaningful pharmacological description is non-trivial; thus we explicitly state in the LLM prompt that the task is a black-box function, and the task description $a$ is set to an empty string.

The \texttt{median\_1} task aims to generate molecules with intermediate structures between menthol and camphor; therefore, the effects of both are listed.
The \texttt{median\_2} task aims to generate molecules with intermediate structures between tadalafil and sildenafil, and both have the same effect.
\begin{table}
  \caption{Hyperparameters.}
  \label{tab:hparams}
  \resizebox{0.9\columnwidth}{!}{%
  \begin{tabular}{l|cc}
    \toprule
    \textbf{Parameter} & \molleo & \proposed{} \\
    & &  and GP BO \\
    \midrule
    Initial samples $N_\mathrm{init}$ & 10 & 10 \\
    Population size $N_\mathrm{elite}$ & 30 & 30 \\
    Number of pairs $N_\mathrm{pairs}$ & 10 & 10 \\
    Number of siblings per pair $N_\mathrm{siblings}$ & 1 & 5 \\
    Batch size $N_\mathrm{batch}$ & 10 & 1 \\
    Candidate pool size $N_\mathrm{cand}$ & 0 & 300 \\
    \bottomrule
  \end{tabular}
  }
\end{table}

\begin{lstlisting}[float=t,caption={Prompt template used in \funcname{EditMol} subroutine for \proposedl{} and \molleo.}, label=lst:prompt_template, escapechar=@, basicstyle=\ttfamily\scriptsize]
system_prompt = "You are an expert medicinal chemist collaborating on molecular design. Given scored exemplars, design a new molecule that could outperform them."

user_prompt_template = """
{
  "description": "You are given two candidate molecules and their scores from an optimization oracle derived from multiple medicinal chemistry heuristics. {TASK_DESC}",
  "candidates": [
    {
      "smiles": "{PARENT1_SMILES}",
      "score": {PARENT1_SCORE}
    },
    {
      "smiles": "{PARENT2_SMILES}",
      "score": {PARENT2_SCORE}
    }
  ],
  "instructions": "Please propose a chemically valid, synthesizable molecule that should achieve a higher desirability score while balancing exploration of new chemotypes with refinement around promising motifs. Generate {NUM_SIBLINGS} distinct candidate molecules.  Each entry must use SMILES and all candidates must be returned in the 'smiles' field.   Respond with a single JSON object containing only the key 'smiles' mapped to an array of exactly {NUM_SIBLINGS} unique SMILES strings. Do not include any additional narrative text."
}
"""
\end{lstlisting}

\begin{table*}[t]
  \centering
  \scriptsize
  \setlength{\tabcolsep}{2pt}
  \caption{Task descriptions for LLM prompt templates and CLAMP text inputs.}
  \label{tab:task-objectives}
  \begin{tabular}{p{0.17\textwidth}p{0.41\textwidth}p{0.38\textwidth}}
    \toprule
     & \textbf{\texttt{\{TASK\_DESC\}} in List.~\ref{lst:prompt_template}} & \textbf{Task description for CLAMP: $a$} \\
    \midrule
    \multicolumn{3}{l}{\textbf{Drug effect optimization}} \\
    \midrule
    albuterol\_similarity & The oracle rewards molecules that act as beta-2 adrenergic receptor agonists. & Beta-2 adrenergic receptor agonists \\
    amlodipine\_mpo & The oracle rewards molecules that inhibit the L-type calcium channel. & Inhibition of the L-type calcium channel \\
    celecoxib\_rediscovery & The oracle rewards molecules that inhibit cyclooxygenase-2. & Inhibition of cyclooxygenase-2 \\
    drd2\_docking & The oracle rewards molecules that act as dopamine D2 receptor ligands. & Dopamine D2 receptor ligand \\
    fexofenadine\_mpo & The oracle rewards molecules that inhibit the histamine H1 receptor. & Inhibition of the histamine H1 receptor \\
    gsk3\_beta & The oracle rewards molecules that inhibit glycogen synthase kinase-3-beta. & Inhibition of glycogen synthase kinase-3-beta \\
    jnk3 & The oracle rewards molecules that inhibit c-Jun N-terminal kinase 3. & Inhibition of c-Jun N-terminal kinase 3 \\
    median\_1 & This oracle rewards molecules that act as TRPV3/TRPM8 agonists. & Agonism of human TRPV3/TRPM8 \\
    median\_2 & This oracle rewards molecules that inhibit PDE5A catalytic activity. & Inhibition of recombinant human PDE5A catalytic activity \\
    mestranol\_similarity & The oracle rewards molecules that inhibit estrogen receptor binding activity. & Inhibition of estrogen receptor binding activity \\
    osimetrinib\_mpo & The oracle rewards molecules that inhibit EGFR tyrosine kinase activity. & Inhibition of EGFR tyrosine kinase activity \\
    perindopril\_mpo & The oracle rewards molecules that inhibit the angiotensin-converting enzyme. & Inhibition of the angiotensin-converting enzyme \\
    ranolazine\_mpo & The oracle rewards molecules that inhibit late sodium current. & Inhibition of late sodium current \\
    sitagliptin\_mpo & The oracle rewards molecules that inhibit dipeptidyl peptidase 4 enzyme activity. & Inhibition of dipeptidyl peptidase 4 enzyme activity \\
    thiothixene\_rediscovery & The oracle rewards molecules that act as dopamine D2 receptor antagonists. & Dopamine D2 receptor antagonists \\
    troglitazone\_rediscovery & The oracle rewards molecules that act as human peroxisome proliferator-activated receptor gamma agonists. & Human peroxisome proliferator-activated receptor gamma agonists \\
    zaleplon\_mpo & The oracle rewards molecules that act as Gamma-Aminobutyric Acid A receptor agonists. & Gamma-Aminobutyric Acid A receptor agonists \\
    \midrule
    \multicolumn{3}{l}{\textbf{Property optimization}} \\
    \midrule
    rdkit\_qed & The oracle returns the Quantitative Estimate of Drug-likeness (QED) normalized to [0,1]; higher values indicate a more drug-like profile. External tool invocation is prohibited. & Quantitative Estimate of Drug-likeness (QED) \\
    \midrule
    \multicolumn{3}{l}{\textbf{Structure-based optimization}} \\
    \midrule
    deco\_hop & \multirow{4}{0.41\textwidth}{This oracle is a black-box function; its analytical form is unavailable and it can only be probed via oracle evaluations.} & \multirow{4}{0.41\textwidth}{N/A (input empty string)} \\
    scaffold\_hop &  &  \\
    isomer\_c7h8n2o2 &  &  \\
    isomer\_c9h10n2o2pf2cl &  &  \\
    \bottomrule
  \end{tabular}
\end{table*}

\section{Additional Experimental Results} \label{sec:appendix_additional_results}
\subsection{Additional analysis of critic contributions}
\begin{figure}
  \centering
  \includegraphics[width=0.45\textwidth]{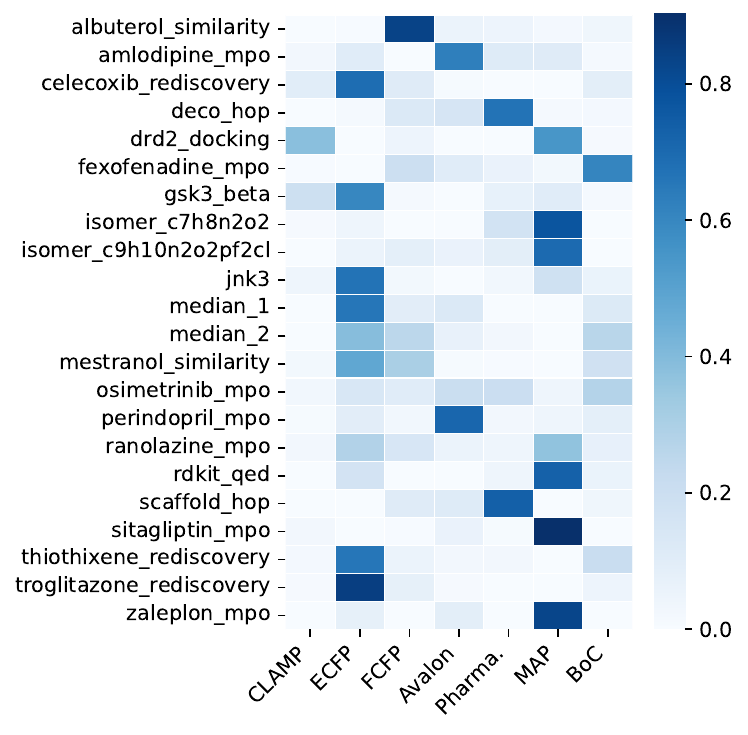}
  \caption{
    Heatmap of critic contributions during the optimization process of \proposedg{}.
    Compared to \proposedl{} shown in Figure~\ref{fig:modal_contrib_mollibral}, 
    the dominant critics for each task are generally consistent.
  }
  \label{fig:modal_contrib_mollibrag}
\end{figure}

\begin{figure}
  \centering
  \includegraphics[width=0.3\textwidth]{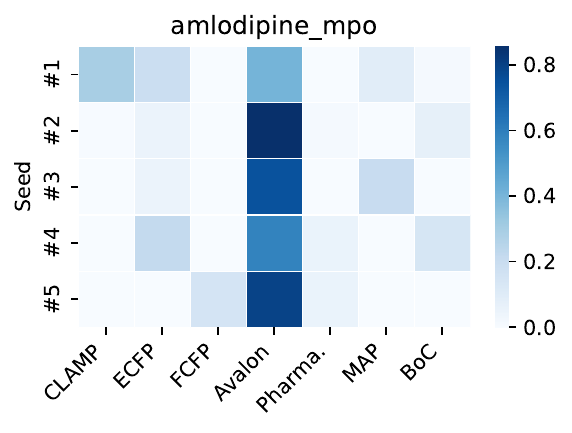}
  \includegraphics[width=0.3\textwidth]{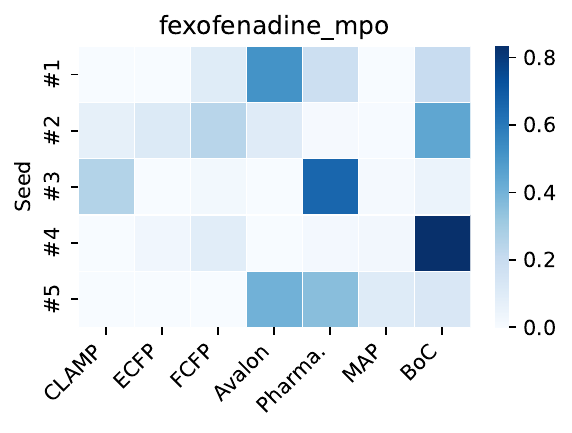}
  \caption{ 
    Heatmaps showing the critic contributions for different seeds.
    While amlodipine\_mpo shows consistent critic contributions across seeds, fexofenadine\_mpo exhibits large variation depending on the seed.
  }
  \label{fig:modal_contrib_mollibral_seeds}
\end{figure}

Figure~\ref{fig:modal_contrib_mollibrag} shows the critic contributions during the optimization process of \proposedg{}.
Figure~\ref{fig:modal_contrib_mollibral_seeds} presents heatmaps of critic contributions for different random seeds in two tasks.

\subsection{Comparison with Latent-space Optimization Methods}
Table~\ref{tab:cowboys} presents the Top-10 AUC comparison under the PMO-300 budget ($N=300$ oracle calls) with latent-space optimization methods including COWBOYS~\citep{Moss2025COWBOYS}.
GA (Genetic Algorithm), CMA-ES, Random Line BO, and TURBO are general optimization methods, and they have been evaluated as applications to molecular optimization in latent space by \citet{GonzalezDuque2024HDBO}.
All latent-space optimization results in Table~\ref{tab:cowboys} are evaluated using the COWBOYS codebase%
\footnote{\url{https://github.com/henrymoss/ROTLSC/tree/main}}.
Although the prior latent-space studies report effectiveness on PMO-300, \proposed{} outperforms latent-space optimization methods even in this low-budget setting.
\begin{table*}[t]
\centering
\setlength{\tabcolsep}{6pt}
\renewcommand{\arraystretch}{1.1}
\caption{
  Top-10 AUC comparison on PMO-300 with latent-space optimization methods in the low budget regime.
} \label{tab:cowboys}
\resizebox{0.85\textwidth}{!}{%
\begin{tabular}{l|ccccccc}
\toprule
& GA & CMA-ES & Random Line BO & TURBO & COWBOYS & \proposedg & \proposedl \\
& \multicolumn{4}{c}{\footnotesize{\citep{GonzalezDuque2024HDBO}}} & \footnotesize{\citep{Moss2025COWBOYS}} & (ours) & (ours) \\
\midrule
Latent-space optimization & $\checkmark$ & $\checkmark$ & $\checkmark$ & $\checkmark$ & $\checkmark$ &  &  \\
\midrule
albuterol\_similarity & 0.321 \pms{0.024} & 0.330 \pms{0.041} & 0.347 \pms{0.036} & 0.357 \pms{0.040} & 0.357 \pms{0.034} & \secondcell{0.748 \pms{0.021}} & \bestcell{0.890 \pms{0.043}} \\
amlodipine\_mpo & 0.426 \pms{0.014} & 0.386 \pms{0.009} & 0.387 \pms{0.010} & 0.374 \pms{0.015} & 0.392 \pms{0.013} & \secondcell{0.488 \pms{0.041}} & \bestcell{0.508 \pms{0.045}} \\
celecoxib\_rediscovery & 0.197 \pms{0.006} & 0.176 \pms{0.007} & 0.176 \pms{0.009} & 0.167 \pms{0.006} & 0.192 \pms{0.019} & \secondcell{0.366 \pms{0.026}} & \bestcell{0.588 \pms{0.073}} \\
deco\_hop & 0.557 \pms{0.006} & 0.538 \pms{0.004} & 0.541 \pms{0.009} & 0.536 \pms{0.003} & 0.546 \pms{0.010} & \secondcell{0.591 \pms{0.010}} & \bestcell{0.597 \pms{0.012}} \\
drd2\_docking & 0.064 \pms{0.082} & 0.034 \pms{0.015} & 0.095 \pms{0.064} & 0.051 \pms{0.023} & 0.079 \pms{0.082} & \secondcell{0.669 \pms{0.294}} & \bestcell{0.912 \pms{0.029}} \\
fexofenadine\_mpo & 0.544 \pms{0.036} & 0.585 \pms{0.027} & 0.612 \pms{0.019} & 0.594 \pms{0.014} & 0.598 \pms{0.016} & \secondcell{0.689 \pms{0.039}} & \bestcell{0.689 \pms{0.040}} \\
gsk3\_beta & 0.159 \pms{0.025} & 0.179 \pms{0.016} & 0.238 \pms{0.073} & 0.160 \pms{0.016} & 0.222 \pms{0.021} & \secondcell{0.333 \pms{0.054}} & \bestcell{0.564 \pms{0.050}} \\
isomer\_c7h8n2o2 & 0.592 \pms{0.106} & 0.601 \pms{0.066} & 0.578 \pms{0.070} & 0.517 \pms{0.079} & 0.666 \pms{0.081} & \bestcell{0.751 \pms{0.015}} & \secondcell{0.713 \pms{0.069}} \\
isomer\_c9h10n2o2pf2cl & 0.477 \pms{0.023} & 0.401 \pms{0.061} & 0.470 \pms{0.095} & 0.444 \pms{0.055} & 0.444 \pms{0.054} & \bestcell{0.653 \pms{0.108}} & \secondcell{0.564 \pms{0.270}} \\
jnk3 & 0.085 \pms{0.009} & 0.074 \pms{0.009} & 0.077 \pms{0.011} & 0.080 \pms{0.010} & 0.083 \pms{0.008} & \secondcell{0.234 \pms{0.070}} & \bestcell{0.297 \pms{0.107}} \\
median\_1 & 0.131 \pms{0.014} & 0.121 \pms{0.004} & 0.124 \pms{0.019} & 0.121 \pms{0.005} & 0.162 \pms{0.032} & \bestcell{0.279 \pms{0.011}} & \secondcell{0.225 \pms{0.030}} \\
median\_2 & 0.142 \pms{0.011} & 0.126 \pms{0.004} & 0.126 \pms{0.004} & 0.124 \pms{0.006} & 0.135 \pms{0.008} & \secondcell{0.193 \pms{0.026}} & \bestcell{0.230 \pms{0.018}} \\
mestranol\_similarity & 0.250 \pms{0.031} & 0.287 \pms{0.022} & 0.299 \pms{0.020} & 0.302 \pms{0.028} & 0.330 \pms{0.016} & \bestcell{0.477 \pms{0.084}} & \secondcell{0.470 \pms{0.020}} \\
osimetrinib\_mpo & 0.672 \pms{0.036} & 0.597 \pms{0.027} & 0.596 \pms{0.018} & 0.600 \pms{0.041} & 0.639 \pms{0.007} & \secondcell{0.708 \pms{0.024}} & \bestcell{0.750 \pms{0.039}} \\
perindopril\_mpo & 0.321 \pms{0.110} & 0.229 \pms{0.031} & 0.265 \pms{0.037} & 0.235 \pms{0.053} & 0.255 \pms{0.033} & \secondcell{0.455 \pms{0.026}} & \bestcell{0.483 \pms{0.024}} \\
ranolazine\_mpo & 0.299 \pms{0.081} & 0.420 \pms{0.026} & 0.378 \pms{0.129} & 0.378 \pms{0.053} & 0.481 \pms{0.030} & \secondcell{0.653 \pms{0.012}} & \bestcell{0.689 \pms{0.029}} \\
rdkit\_qed & 0.884 \pms{0.031} & 0.823 \pms{0.029} & 0.820 \pms{0.037} & 0.816 \pms{0.019} & 0.817 \pms{0.021} & \secondcell{0.903 \pms{0.014}} & \bestcell{0.923 \pms{0.006}} \\
scaffold\_hop & 0.421 \pms{0.009} & 0.397 \pms{0.017} & 0.406 \pms{0.014} & 0.392 \pms{0.007} & 0.410 \pms{0.007} & \secondcell{0.483 \pms{0.002}} & \bestcell{0.484 \pms{0.013}} \\
sitagliptin\_mpo & 0.119 \pms{0.071} & 0.099 \pms{0.034} & 0.152 \pms{0.066} & 0.074 \pms{0.013} & 0.186 \pms{0.067} & \secondcell{0.236 \pms{0.063}} & \bestcell{0.298 \pms{0.056}} \\
thiothixene\_rediscovery & 0.218 \pms{0.014} & 0.195 \pms{0.008} & 0.195 \pms{0.005} & 0.194 \pms{0.008} & 0.200 \pms{0.019} & \secondcell{0.326 \pms{0.036}} & \bestcell{0.480 \pms{0.079}} \\
troglitazone\_rediscovery & 0.181 \pms{0.016} & 0.160 \pms{0.007} & 0.158 \pms{0.008} & 0.160 \pms{0.003} & 0.170 \pms{0.006} & \bestcell{0.316 \pms{0.031}} & \secondcell{0.310 \pms{0.030}} \\
zaleplon\_mpo & 0.278 \pms{0.087} & 0.225 \pms{0.021} & 0.283 \pms{0.037} & 0.216 \pms{0.044} & 0.259 \pms{0.052} & \secondcell{0.386 \pms{0.026}} & \bestcell{0.394 \pms{0.053}} \\
\midrule
Sum & 7.339 & 6.983 & 7.319 & 6.892 & 7.621 & \secondcell{10.934} & \bestcell{12.058} \\
\bottomrule
\end{tabular}
}
\end{table*}

\subsection{Additional Ablation Study}
\paragraph{Ensemble method for GPs}
This section presents an ablation study on the GP ensemble method.
Table~\ref{tab:poe} shows the comparison with Product of Experts (PoE)~\citep{cao2014generalized} as an alternative to the selection-based ensemble method~\citep{lu2023surrogate} that we use in the main text.
Comparing the Sum row, the selection method outperforms PoE.
The difference is small but consistent in both \proposedg{} and \proposedl{}; therefore, we adopted the selection method in the main text.
A possible reason for the similar results is that both the selection method and PoE assign dynamic weights to the GPs. 
PoE takes the product of the predictive distributions of each GP (i.e., $\prod^{M}_{m=1} \mathcal{N} (f; \mu_{m}, \sigma^{2}_{m})$) as the ensemble result, which is equivalent to a weighted average of the predictive means $\{\mu_{m}\}$ with weights based on the precisions $\{\sigma^{-2}_{m}\}$.
\begin{table}[t]
\centering
\caption{
  Ablation study results on the GP ensemble method (Top-10 AUC on PMO-300), comparing the selection method~\citep{lu2023surrogate} used in the main text and PoE~\citep{cao2014generalized}.
  }
\label{tab:poe}
\resizebox{\columnwidth}{!}{%
\begin{tabular}{l|cc|cc}
\toprule
 & \multicolumn{2}{c}{\proposedg} & \multicolumn{2}{|c}{\proposedl} \\
 \midrule
 GP ensemble & Selection & PoE & Selection & PoE \\
\midrule
albuterol\_similarity & \bestcell{0.748 \pms{0.021}} & \secondcell{0.741 \pms{0.085}} & \secondcell{0.890 \pms{0.043}} & \bestcell{0.900 \pms{0.055}} \\
amlodipine\_mpo & \secondcell{0.488 \pms{0.041}} & \bestcell{0.496 \pms{0.053}} & \secondcell{0.508 \pms{0.045}} & \bestcell{0.527 \pms{0.049}} \\
celecoxib\_rediscovery & \secondcell{0.366 \pms{0.026}} & \bestcell{0.378 \pms{0.021}} & \bestcell{0.588 \pms{0.073}} & \secondcell{0.484 \pms{0.059}} \\
deco\_hop & \secondcell{0.591 \pms{0.010}} & \bestcell{0.593 \pms{0.011}} & \secondcell{0.597 \pms{0.012}} & \bestcell{0.601 \pms{0.012}} \\
drd2\_docking & \bestcell{0.669 \pms{0.294}} & \secondcell{0.643 \pms{0.290}} & \bestcell{0.912 \pms{0.029}} & \secondcell{0.876 \pms{0.038}} \\
fexofenadine\_mpo & \bestcell{0.689 \pms{0.039}} & \secondcell{0.664 \pms{0.066}} & \bestcell{0.689 \pms{0.040}} & \secondcell{0.672 \pms{0.061}} \\
gsk3\_beta & \secondcell{0.333 \pms{0.054}} & \bestcell{0.356 \pms{0.055}} & \secondcell{0.564 \pms{0.050}} & \bestcell{0.571 \pms{0.029}} \\
isomer\_c7h8n2o2 & \bestcell{0.751 \pms{0.015}} & \secondcell{0.725 \pms{0.090}} & \bestcell{0.713 \pms{0.069}} & \secondcell{0.612 \pms{0.089}} \\
isomer\_c9h10n2o2pf2cl & \bestcell{0.653 \pms{0.108}} & \secondcell{0.636 \pms{0.085}} & \secondcell{0.564 \pms{0.270}} & \bestcell{0.676 \pms{0.103}} \\
jnk3 & \bestcell{0.234 \pms{0.070}} & \secondcell{0.222 \pms{0.057}} & \secondcell{0.297 \pms{0.107}} & \bestcell{0.337 \pms{0.034}} \\
median\_1 & \secondcell{0.279 \pms{0.011}} & \bestcell{0.281 \pms{0.013}} & \bestcell{0.225 \pms{0.030}} & \secondcell{0.221 \pms{0.049}} \\
median\_2 & \bestcell{0.193 \pms{0.026}} & \secondcell{0.192 \pms{0.021}} & \bestcell{0.230 \pms{0.018}} & \secondcell{0.218 \pms{0.017}} \\
mestranol\_similarity & \bestcell{0.477 \pms{0.084}} & \secondcell{0.414 \pms{0.042}} & \bestcell{0.470 \pms{0.020}} & \secondcell{0.455 \pms{0.045}} \\
osimetrinib\_mpo & \bestcell{0.708 \pms{0.024}} & \secondcell{0.692 \pms{0.048}} & \secondcell{0.750 \pms{0.039}} & \bestcell{0.756 \pms{0.067}} \\
perindopril\_mpo & \bestcell{0.455 \pms{0.026}} & \secondcell{0.442 \pms{0.022}} & \bestcell{0.483 \pms{0.024}} & \secondcell{0.459 \pms{0.012}} \\
ranolazine\_mpo & \secondcell{0.653 \pms{0.012}} & \bestcell{0.656 \pms{0.028}} & \secondcell{0.689 \pms{0.029}} & \bestcell{0.697 \pms{0.015}} \\
rdkit\_qed & \secondcell{0.903 \pms{0.014}} & \bestcell{0.909 \pms{0.014}} & \bestcell{0.923 \pms{0.006}} & \secondcell{0.915 \pms{0.017}} \\
scaffold\_hop & \secondcell{0.483 \pms{0.002}} & \bestcell{0.492 \pms{0.012}} & \bestcell{0.484 \pms{0.013}} & \secondcell{0.483 \pms{0.016}} \\
sitagliptin\_mpo & \bestcell{0.236 \pms{0.063}} & \secondcell{0.213 \pms{0.078}} & \secondcell{0.298 \pms{0.056}} & \bestcell{0.322 \pms{0.087}} \\
thiothixene\_rediscovery & \secondcell{0.326 \pms{0.036}} & \bestcell{0.370 \pms{0.037}} & \bestcell{0.480 \pms{0.079}} & \secondcell{0.430 \pms{0.034}} \\
troglitazone\_rediscovery & \bestcell{0.316 \pms{0.031}} & \secondcell{0.309 \pms{0.020}} & \bestcell{0.310 \pms{0.030}} & \secondcell{0.301 \pms{0.038}} \\
zaleplon\_mpo & \bestcell{0.386 \pms{0.026}} & \secondcell{0.368 \pms{0.047}} & \secondcell{0.394 \pms{0.053}} & \bestcell{0.401 \pms{0.068}} \\
\midrule
Sum & \bestcell{10.934} & \secondcell{10.791} & \bestcell{12.058} & \secondcell{11.912} \\
\bottomrule
\end{tabular}
}
\end{table}

\paragraph{Gating method for CLAMP}
This section presents an ablation study on the gating method for the CLAMP critic.
Table~\ref{tab:gating_ablation} shows the comparison results with
EXP3~\citep{auer2002nonstochastic} and LLIMBO~\citep{chang2025llinbo} as alternatives to the correlation-based gating method that we use.
EXP3 is a multi-armed bandit algorithm, and here it learns the gating probabilities using score improvement as rewards.
Since the improvements are rarely observed, the learning speed tends to be slow, resulting in EXP3 performing the worst in both \proposedg{} and \proposedl{}.
LLIMBO is an algorithm for combining candidates proposed by a GP and  LLM in general black-box optimization formulations.
In this paper, LLIMBO computes the quantile of the CLAMP-proposed candidate among all candidates ranked by the GP critics, and then uses that quantile as the gating probability%
\footnote{
This method is a modification of LLIMBO-Justify proposed in the original paper~\citep{chang2025llinbo}, adapted to probabilistically select whether to adopt CLAMP or not.
}.
Although this method is competitive with the correlation-based method in both \proposedg{} and \proposedl{}, we adopted the  correlation-based method in the main text because of its simplicity.
\begin{table*}[t]
\centering
\setlength{\tabcolsep}{6pt}
\renewcommand{\arraystretch}{1.1}
\caption{
  Ablation study results on the CLAMP gating method (Top-10 AUC on PMO-300), comparing the correlation-based method (denoted as Corr.) used in the main text with LLIMBO~\citep{chang2025llinbo} and EXP3~\citep{auer2002nonstochastic}.
  }
\resizebox{0.7\textwidth}{!}{%
\begin{tabular}{l|ccc|ccc}
\toprule
& \multicolumn{3}{c}{\proposedg{}} & \multicolumn{3}{|c}{\proposedl{}} \\
\midrule
CLAMP gating & Corr. & LLIMBO & EXP3 & Corr. & LLIMBO & EXP3 \\
\midrule
albuterol\_similarity & \bestcell{0.748 \pms{0.021}} & \secondcell{0.745 \pms{0.036}} & 0.718 \pms{0.085} & 0.890 \pms{0.043} & \bestcell{0.912 \pms{0.019}} & \secondcell{0.893 \pms{0.027}} \\
amlodipine\_mpo & \bestcell{0.488 \pms{0.041}} & \secondcell{0.483 \pms{0.059}} & 0.456 \pms{0.051} & 0.508 \pms{0.045} & \bestcell{0.541 \pms{0.039}} & \secondcell{0.520 \pms{0.051}} \\
celecoxib\_rediscovery & \secondcell{0.366 \pms{0.026}} & \bestcell{0.392 \pms{0.052}} & 0.343 \pms{0.015} & \bestcell{0.588 \pms{0.073}} & \secondcell{0.497 \pms{0.069}} & 0.496 \pms{0.102} \\
deco\_hop & 0.591 \pms{0.010} & \bestcell{0.599 \pms{0.010}} & \secondcell{0.597 \pms{0.008}} & 0.597 \pms{0.012} & \bestcell{0.605 \pms{0.019}} & \secondcell{0.604 \pms{0.015}} \\
drd2\_docking & \secondcell{0.669 \pms{0.294}} & \bestcell{0.755 \pms{0.132}} & 0.650 \pms{0.347} & \secondcell{0.912 \pms{0.029}} & 0.911 \pms{0.023} & \bestcell{0.930 \pms{0.015}} \\
fexofenadine\_mpo & \bestcell{0.689 \pms{0.039}} & \secondcell{0.656 \pms{0.042}} & 0.645 \pms{0.052} & \bestcell{0.689 \pms{0.040}} & \secondcell{0.674 \pms{0.051}} & 0.664 \pms{0.047} \\
gsk3\_beta & \secondcell{0.333 \pms{0.054}} & \bestcell{0.352 \pms{0.068}} & 0.315 \pms{0.048} & \bestcell{0.564 \pms{0.050}} & \secondcell{0.556 \pms{0.105}} & 0.551 \pms{0.055} \\
isomer\_c7h8n2o2 & \secondcell{0.751 \pms{0.015}} & \bestcell{0.784 \pms{0.027}} & 0.653 \pms{0.060} & \bestcell{0.713 \pms{0.069}} & \secondcell{0.660 \pms{0.066}} & 0.555 \pms{0.236} \\
isomer\_c9h10n2o2pf2cl & \secondcell{0.653 \pms{0.108}} & \bestcell{0.659 \pms{0.070}} & 0.575 \pms{0.067} & \secondcell{0.564 \pms{0.270}} & \bestcell{0.618 \pms{0.140}} & 0.416 \pms{0.254} \\
jnk3 & \bestcell{0.234 \pms{0.070}} & 0.207 \pms{0.039} & \secondcell{0.225 \pms{0.050}} & 0.297 \pms{0.107} & \bestcell{0.356 \pms{0.092}} & \secondcell{0.324 \pms{0.114}} \\
median\_1 & \bestcell{0.279 \pms{0.011}} & \secondcell{0.272 \pms{0.024}} & 0.255 \pms{0.014} & \bestcell{0.225 \pms{0.030}} & \secondcell{0.213 \pms{0.029}} & 0.196 \pms{0.015} \\
median\_2 & \bestcell{0.193 \pms{0.026}} & \secondcell{0.191 \pms{0.022}} & 0.185 \pms{0.021} & \bestcell{0.230 \pms{0.018}} & \secondcell{0.212 \pms{0.010}} & 0.210 \pms{0.018} \\
mestranol\_similarity & \bestcell{0.477 \pms{0.084}} & 0.469 \pms{0.040} & \secondcell{0.473 \pms{0.052}} & \bestcell{0.470 \pms{0.020}} & \secondcell{0.444 \pms{0.038}} & 0.415 \pms{0.035} \\
osimetrinib\_mpo & \secondcell{0.708 \pms{0.024}} & \bestcell{0.724 \pms{0.018}} & 0.679 \pms{0.057} & \secondcell{0.750 \pms{0.039}} & \bestcell{0.768 \pms{0.029}} & 0.741 \pms{0.045} \\
perindopril\_mpo & \bestcell{0.455 \pms{0.026}} & \secondcell{0.448 \pms{0.032}} & 0.420 \pms{0.014} & \secondcell{0.483 \pms{0.024}} & \bestcell{0.486 \pms{0.016}} & 0.469 \pms{0.024} \\
ranolazine\_mpo & \secondcell{0.653 \pms{0.012}} & \bestcell{0.670 \pms{0.022}} & 0.649 \pms{0.039} & \secondcell{0.689 \pms{0.029}} & \bestcell{0.701 \pms{0.016}} & 0.680 \pms{0.020} \\
rdkit\_qed & \secondcell{0.903 \pms{0.014}} & \bestcell{0.904 \pms{0.015}} & 0.896 \pms{0.018} & \bestcell{0.923 \pms{0.006}} & 0.909 \pms{0.024} & \secondcell{0.922 \pms{0.008}} \\
scaffold\_hop & \secondcell{0.483 \pms{0.002}} & \bestcell{0.484 \pms{0.008}} & 0.474 \pms{0.005} & \secondcell{0.484 \pms{0.013}} & \bestcell{0.487 \pms{0.013}} & 0.475 \pms{0.013} \\
sitagliptin\_mpo & \bestcell{0.236 \pms{0.063}} & 0.149 \pms{0.079} & \secondcell{0.195 \pms{0.064}} & \secondcell{0.298 \pms{0.056}} & \bestcell{0.313 \pms{0.092}} & 0.264 \pms{0.111} \\
thiothixene\_rediscovery & \secondcell{0.326 \pms{0.036}} & \bestcell{0.332 \pms{0.013}} & 0.291 \pms{0.021} & \bestcell{0.480 \pms{0.079}} & \secondcell{0.449 \pms{0.073}} & 0.405 \pms{0.048} \\
troglitazone\_rediscovery & \secondcell{0.316 \pms{0.031}} & 0.302 \pms{0.027} & \bestcell{0.317 \pms{0.029}} & \secondcell{0.310 \pms{0.030}} & \bestcell{0.318 \pms{0.049}} & 0.282 \pms{0.021} \\
zaleplon\_mpo & \bestcell{0.386 \pms{0.026}} & \secondcell{0.383 \pms{0.034}} & 0.362 \pms{0.014} & \bestcell{0.394 \pms{0.053}} & \secondcell{0.384 \pms{0.032}} & 0.380 \pms{0.046} \\
\midrule
Sum & \secondcell{10.934} & \bestcell{10.960} & 10.375 & \bestcell{12.058} & \secondcell{12.015} & 11.391 \\
\bottomrule
\end{tabular}
}
\label{tab:gating_ablation}
\end{table*}

\subsection{Wall-clock Time and API Call Cost}
Table~\ref{tab:cost_comparison} presents the wall-clock time and API call cost comparison on the PMO-1K benchmark.
We run GP training/inference and CLAMP evaluation on an NVIDIA H200 GPU.
Since \proposed{} uses multiple critics, it is computationally more expensive than Tripp's GP BO, which uses a single critic; however, the increase in time per iteration is only a few seconds.
While \proposedl{} incurs approximately $20\times$ the API call cost compared to \molleo{}, the total cost per seed is about USD 5.
%
\begin{table}
  \centering
  \caption{
    Wall-clock time and LLM API (GPT-5-mini) call cost comparison on the PMO-1K benchmark.
    These results are averaged over 5 seeds on the \texttt{albuterol\_similarity} task.
  }
  \resizebox{0.95\columnwidth}{!}{%
  \begin{tabular}{l|cccc}
    \toprule
    \textbf{Method} & \textbf{Total time} & \textbf{Avg. time per iter.} & \textbf{Total API cost} \\
    \midrule
    \proposedg{} & 3.32h & 11.95s & - \\
    \proposedl{} & 4.22h & 15.19s & 4.88 USD \\
    Tripp's GP  & 2.34h & 8.42s & - \\
    \molleo{} & 16.4 min & 0.99s & 0.25 USD \\
    \bottomrule
  \end{tabular}
  }
  \label{tab:cost_comparison}
\end{table}

\end{document}